%% file: iclr2027_conference.tex
\documentclass{article} 
\usepackage{iclr2027_conference,times}

\input{math_commands.tex}

\usepackage{hyperref}
\usepackage{url}
\usepackage{booktabs}
\usepackage{graphicx}
\usepackage{float}
\usepackage{placeins}
\usepackage{multirow}
\usepackage{capt-of}
\usepackage{enumitem}
\usepackage{colortbl}
\usepackage{algorithm}
\usepackage{algpseudocode}
\usepackage[T1]{fontenc}
\usepackage[utf8]{inputenc}
\usepackage{amssymb}
\usepackage[table]{xcolor}
\usepackage{arydshln}
\usepackage{wrapfig}
\usepackage{ragged2e}
\usepackage{xcolor}

\hypersetup{
    hidelinks
}

\definecolor{bestcolor}{HTML}{F1C2C2}
\definecolor{secondcolor}{HTML}{F8E1C1}

\title{TACC: Target-Aware Carrier Compression for Dense-Context Novel View Synthesis}

\newcommand{\authorline}[1]{%
    \makebox[\dimexpr\textwidth-2\tabcolsep\relax][c]{#1}%
}

\author{%
\authorline{\bfseries
Yiming Fan$^{1,*}$ \enspace
Yijia Guo$^{2,*,\ddagger}$ \enspace
Zewen Guo$^{3}$ \enspace
Liwen Hu$^{2}$ \enspace
Yuheng Yuan \enspace
Yang Li$^{2}$} \\[3pt]
\authorline{\bfseries
Zhiwei Li$^{4}$ \enspace
Xitong Ling$^{5}$ \enspace
Caihong Yuan$^{1}$ \enspace
Shengbo Chen$^{4,\dagger}$ \enspace
Tiejun Huang$^{2}$ \enspace
Lei Ma$^{2,\dagger}$} \\[6pt]
\authorline{\normalfont
$^{1}$Henan University \quad
$^{2}$Peking University \quad
$^{3}$New York University} \\[2pt]
\authorline{\normalfont
$^{4}$Nanchang University \quad
$^{5}$Tsinghua University}%
}

\iclrfinalcopy

\begin{document}

\maketitle

\fancyhead{}
\renewcommand{\headrulewidth}{0pt}


\begingroup
\renewcommand{\thefootnote}{}
\footnotetext[0]{%
    \begin{tabular}[t]{@{}l@{}}
        $^{*}$Equal contribution.\quad
        $^{\dagger}$Corresponding authors:
        chenshengbo@ncu.edu.cn; lei.ma@pku.edu.cn.\\
        $^{\ddagger}$Project leader.
    \end{tabular}%
}
\endgroup

\begin{figure}[H]

    \centering
    \includegraphics[width=\linewidth]{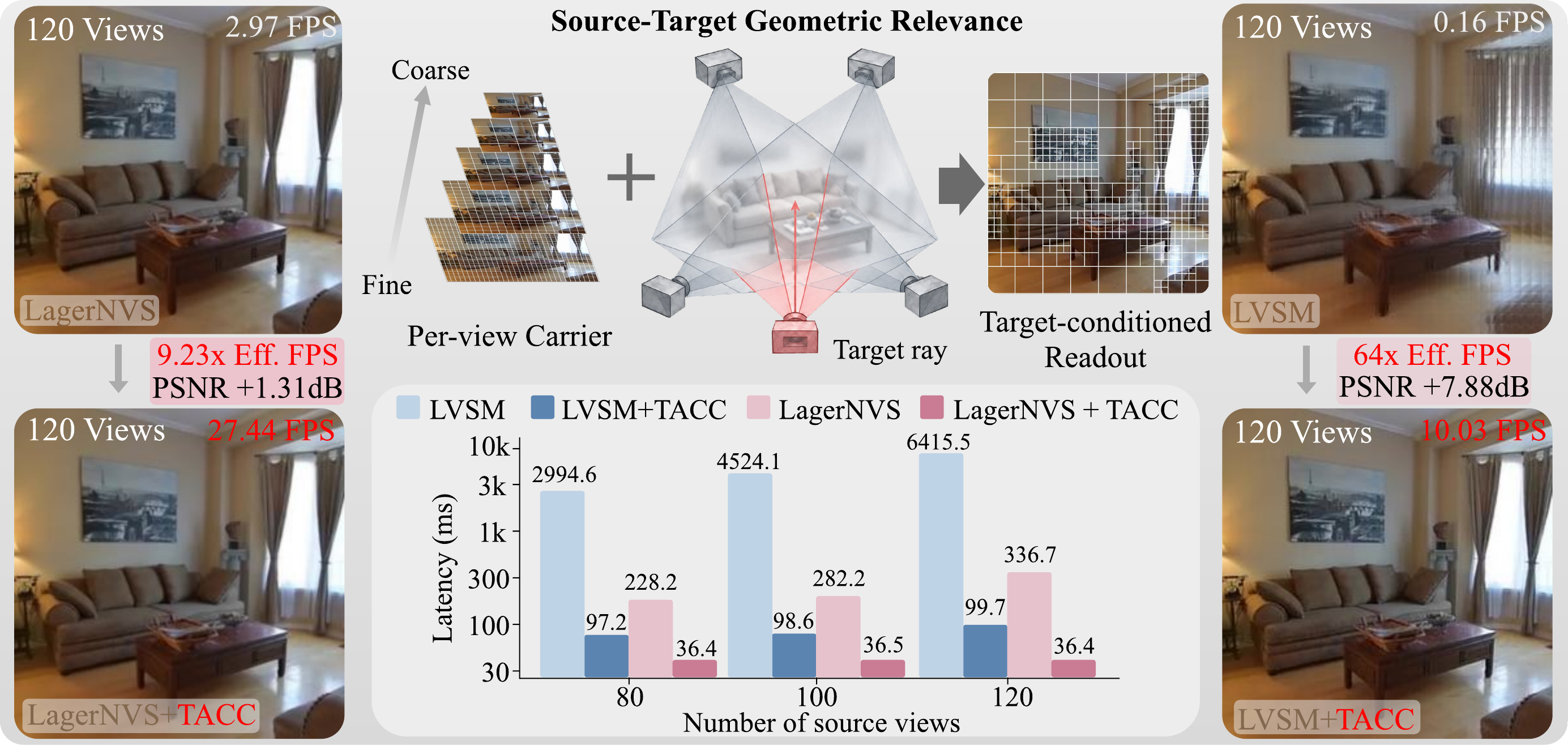}

    \caption{
        TACC organizes source representations into reusable per-view
        Carrier hierarchies for target-conditioned readout.
        On RE10K with 120 source views, TACC achieves $9.23\times$
        and $64\times$ higher Effective FPS than
        frozen Full Context on LagerNVS and LVSM,
        respectively.
    }
    \label{fig:trace_overview}
\end{figure}

\input{sections/abstract}

\input{sections/1_intro}

\input{sections/2_related}

\input{sections/3_method}

\input{sections/4_experiments}

\input{sections/5_conclusion}

\input{sections/6_statements}

\bibliography{iclr2027_conference}
\bibliographystyle{iclr2027_conference}

\appendix
\input{sections/appendix}

\end{document}

%% file: math_commands.tex
\usepackage{amsmath,amsfonts,bm}

\def\eqref#1{equation~\ref{#1}}

\def\1{\bm{1}}

\DeclareMathAlphabet{\mathsfit}{\encodingdefault}{\sfdefault}{m}{sl}
\SetMathAlphabet{\mathsfit}{bold}{\encodingdefault}{\sfdefault}{bx}{n}



%% file: sections/abstract.tex
\begin{abstract}
Recent feed-forward approaches mark a paradigm shift in novel view synthesis by eliminating the reliance on per-scene optimization, directly inferring 3D representations from input views. However, their reliance on global scene-token context incurs substantial
memory and latency overhead, constraining their ability to handle increasingly
dense source-view contexts. Compressing source representations offers an intuitive way to limit this token budget. Prior general-purpose compression methods are not tailored for novel view synthesis, as their fixed, target-agnostic nature fails to adapt evidence allocation to individual target views. To overcome this, we introduce TACC, a target-aware compression module tailored for feed-forward latent NVS. Specifically, TACC organizes each source view into a progressive carrier hierarchy. Driven by source-target geometric relevance, it adaptively retrieves tokens to dynamically allocate exact computational quotas. This selective retrieval directly yields a compact, target-conditioned representation. Extensive experiments on the RE10K demonstrate that under dense input settings, TACC achieves $9.23\times$ and $64\times$ higher Effective FPS than Full Context on LagerNVS and LVSM, respectively. Effective FPS excludes one-time source encoding and Carrier construction.
\end{abstract}

%% file: sections/1_intro.tex
\section{Introduction}
\label{sec:introduction}
Novel view synthesis (NVS) is a fundamental task in computer vision and graphics that has long been dominated by handcrafted 3D representations and slow, per-scene optimization ~\citep{nerf,3dgs}. Recently, feed-forward latent methods ~\citep{rayzer,lvsm} have driven a paradigm shift by completely bypassing explicitly handcrafted 3D reconstructions. These purely transformer-based designs leverage scaled geometry transformers to output new views from tokens directly. By minimizing 3D inductive biases, they transform NVS into a fully data-driven problem. However, this reliance on a global scene-token context creates a fundamental bottleneck in dense source-view settings. Because the rendering pathway must process all source tokens, adapting to dense inputs inevitably inflates the token budget, making the synthesis process computationally prohibitive and causing severe memory and latency overheads~\citep{lvsm,efficientlvsm}.

Compressing the source token space offers the most direct strategy to mitigate this computational bottleneck. However, dedicated compression schemes for latent feed-forward NVS architectures remain largely unexplored. Adapting context reduction strategies from feed-forward 3D reconstruction or Gaussian splatting frameworks is fundamentally inadequate~\citep{fastvggt,come,zpressor}. While these approaches successfully achieve global token reduction, they fail to address how a limited token budget should be allocated across source views. Since varying target cameras rely on different source-view overlaps and visibilities, uniform compression introduces an unavoidable tension between maintaining broad multi-view coverage and preserving high-frequency visual evidence from specific, target-relevant views. Fully unlocking the potential of feed-forward latent NVS requires a compression paradigm that is inherently target-aware.

We fulfill this requirement by introducing \textbf{TACC} (\textbf{T}arget-\textbf{A}ware \textbf{C}arrier \textbf{C}ompression), a novel compression module explicitly designed for feed-forward latent NVS. Specifically, it pre-organizes each source-view representation into a progressive, reusable carrier hierarchy. When queried with a target camera, TACC evaluates source-target geometric relevance to dynamically allocate an exact computational quota across the source views. Guided by these, it retrieves the corresponding token subsets from the hierarchy to directly construct a highly compact, target-conditioned representation. Crucially, only the multi-granularity Carrier module is optimized during training, leaving the pretrained backbone frozen.
Extensive evaluations across three prominent latent NVS backbones~\citep{lagernvs,lvsm,clift} on the DL3DV~\citep{dl3dv} and RealEstate10K~\citep{realestate10k} datasets demonstrate that TACC consistently delivers superior reconstruction quality compared to matched-budget compression baselines. 
As shown in Fig.~\ref{fig:trace_overview}, at 120 source views
on RealEstate10K, TACC achieves $9.23\times$ and
$64\times$ higher Effective FPS than frozen Full Context with LagerNVS and LVSM, respectively,
demonstrating its exceptional scalability and
real-time potential. These target-time speedups
exclude one-time source encoding and Carrier
construction. Our core contributions are summarized as follows:
\begin{itemize}
    \item We introduce TACC, a target-aware compression module for dense-context feed-forward latent NVS that dynamically controls the decoder-visible token budget while keeping pretrained rendering pathways completely unchanged.
    \item We propose a reusable, progressive per-view carrier hierarchy. This structure partitions the source token grid into reading frontiers at every valid positive quota, enabling exact, target-conditioned allocation and readout without redundant re-encoding.
    \item We demonstrate that TACC consistently improves reconstruction quality over matched-budget baselines across three NVS backbones and two datasets, maintaining highly stable target-time throughput even as the source context grows significantly.
\end{itemize}

%% file: sections/2_related.tex
\section{Related Work}

\subsection{Generalizable Novel View Synthesis}

Neural scene representations have enabled diverse applications,
including spike-based novel view synthesis~\citep{Guo2024SpikeNeRF},
real-time relighting~\citep{Guo2024PRTGS}, and large-scale online
reconstruction from multi-camera streams~\citep{Guo2025MultiCameraRigs}.
Generalizable novel view synthesis has evolved from learned image-based
and radiance-field rendering~\citep{ibrnet,mvsnerf,geonerf} to recent
feed-forward models that infer scene representations directly from source
observations~\citep{pixelsplat,mvsplat,gslrm,grm}.
Transformer-based methods further encode multi-view observations into learned
latent representations or process source and target tokens within learned
rendering architectures~\citep{srt,viewformer,lvsm,rayzer,lagernvs}.
Recent geometry-conditioned NVS methods further leverage projected visual
representations in ReNoV~\citep{kwak2026renov}, target-view projective
conditioning in PVSM~\citep{wu2026projections}, and confidence-aware
geometry routing in VGGT-Diff~\citep{chen2026vggtdiff} to guide novel
view synthesis. Tok-D~\citep{nair2025tokd} applies separate modulation to source and
target tokens and explores synthetic multi-view data augmentation
for cross-dataset generalization.
As source-view contexts grow, however, the computation required to process
these representations also increases.
Efficient-LVSM reduces this cost by decoupling source and target
processing~\citep{efficientlvsm}, whereas TACC compresses the source
representation while retaining the pretrained rendering pathway.

\subsection{Visual Token Reduction and Structure Preservation}

As visual token sequences grow longer, their redundancy becomes increasingly
pronounced, motivating pruning, sampling, merging, and learned token
condensation across vision and multimodal
Transformers~\citep{dynamicvit,evit,tokenlearner,avit,ats,tome,divprune}.
These approaches reduce token computation through selection, pruning, merging,
or condensation while preserving task-relevant visual information.
Feed-forward visual geometry Transformers such as VGGT jointly process
multiple views to infer camera and dense 3D attributes~\citep{vggt}.
Recent work therefore extends token reduction to this setting, where preserving
spatial and multi-view structure becomes increasingly important
~\citep{fastvggt,good_token_hunting}.
Co-Me, for example, uses confidence-guided token merging to accelerate visual
geometry Transformers while preserving spatial coverage~\citep{come}.
These methods primarily reduce token computation within visual geometry
backbones; TACC instead controls the source representation exposed to the
target-conditioned rendering pathway under a fixed decoder-visible budget.

\subsection{Compact Multi-View Scene Representations for Neural Rendering}

More directly related to our setting, ZPressor compresses support-view
information into anchor-view features to form a compact latent state for
feed-forward 3D Gaussian Splatting~\citep{zpressor}.
CLiFT constructs compressed light-field tokens and retrieves a target-dependent
subset according to the available rendering budget~\citep{clift}.
These methods establish compact scene representations for efficient neural
rendering, but they do not provide a reusable per-view hierarchy that supports
exact target-conditioned budget allocation across source views.
TACC instead retains a reusable progressive Carrier hierarchy for each source
view and allocates exact integer quotas across views according to target
relevance before reading the corresponding frontiers under a fixed
decoder-visible budget.

%% file: sections/3_method.tex
\section{Method}
\label{sec:method}

\subsection{Overview}
\label{sec:overview}

\begin{figure*}[t]
    \centering
    \includegraphics[width=\textwidth]{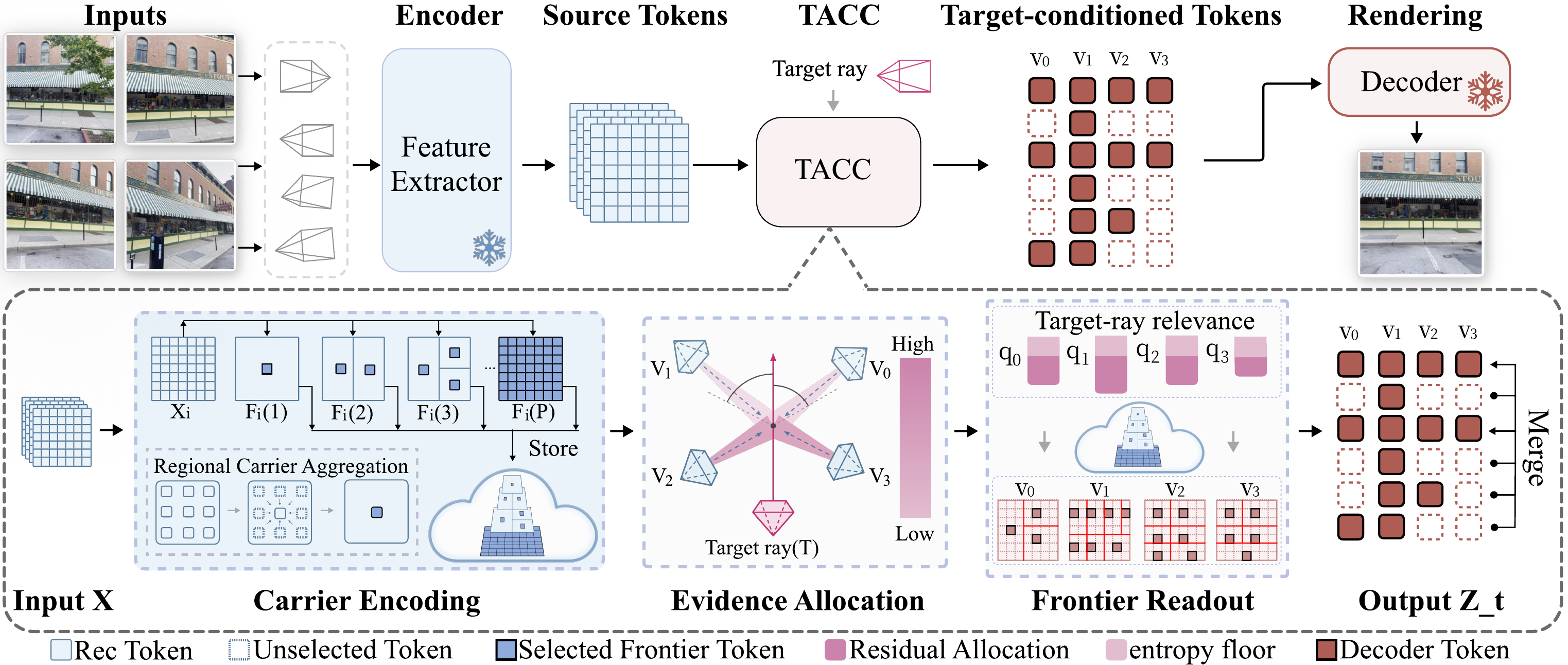}
    \vspace{-15pt}
    \caption{
        \textbf{Overview of TACC for feed-forward latent NVS.}
        Source representations are organized into reusable progressive Carrier
        hierarchies, from which target-aware allocation reads per-view frontiers
        under a fixed decoder-visible budget to form $Z_t$.
    }
    \label{fig:framework}
    \vspace{-0.5cm}
\end{figure*}

Given $V$ source views, TACC operates on a backbone-specific source
representation $X=\{X_i\}_{i=1}^{V}$, where
$X_i=\{x_{i,p}\}_{p=1}^{P}\in\mathbb{R}^{P\times C}$ denotes the
per-view token representation at the TACC insertion point, with $P$ tokens
and feature dimension $C$. The exact construction of $X_i$ depends on the
backbone and is detailed in Appendix~\ref{app:backbone_integration}.
Under a fixed decoder-visible token budget $B=A\cdot P$, TACC compresses
$X$ into a compact target-conditioned representation $Z_t$ containing
$B$ tokens, where $A$ denotes the equivalent view count, i.e., the budget
measured in units of one complete $P$-token source representation, with
$1<A\leq V$.
Let $R_s$ and $R_t$ denote the source-view and target-view camera-ray inputs,
respectively. Figure~\ref{fig:framework} illustrates the overall TACC
pipeline, which is formulated as
\begin{equation}
Z_t
=
\operatorname{TACC}(X,R_s,R_t,B),
\qquad
I_t
=
\mathcal{D}(Z_t,R_t),
\label{eq:tacc_overview}
\end{equation}
where $D$ is the backbone's unchanged target-conditioned rendering pathway.

TACC follows an \emph{encode once, allocate on demand} design. Before the
target view is known, each source representation $X_i$ is organized into a
progressive Carrier hierarchy $H_i=G_\theta(X_i)$. Since the hierarchy
depends only on the fixed source representation, it is constructed once and
reused across target views. Once a target view $t$ becomes available, TACC
allocates the fixed budget $B$ across source views according to their
geometric relevance to the target, reads the corresponding frontier from each
hierarchy, and concatenates the selected Carriers to form $Z_t$.
Progressive Carrier Encoding provides reusable multi-granularity source
representations, while Target-Conditioned Evidence Allocation determines the
per-view quotas and corresponding frontier readout for each target.

\subsection{Progressive Carrier Encoding}
\label{sec:carrier_encoding}

For each per-view source representation $X_i$, TACC constructs a progressive
Carrier hierarchy that supports readout under different per-view token budgets.

\paragraph{Regional Carrier construction.}
Let $X_i=\{x_{i,p}\}_{p=1}^{P}$ denote the source tokens arranged on an
$H\times W$ spatial grid, where $P=HW$.
For a rectangular region $R$, we retain the original token nearest to its
geometric center as the anchor $a_R$; for a single-token region, the anchor
is simply its unique token.
We use a learned residual to encode complementary regional information.

We compute the regional mean
$\mu_R=|R|^{-1}\sum_{p\in R}x_{i,p}$ and the channel-normalized variance
$v_R=(|R|C)^{-1}\sum_{p\in R}\|x_{i,p}-\mu_R\|_2^2$, where $C$ denotes
the feature dimension.
For region $R$, let $h_R$ and $w_R$ denote its height and width on the
$H\times W$ token grid, and define the normalized four-dimensional regional
attributes as
\begin{equation}
s_R
=
\left[
\frac{\log|R|}{\kappa_a},
\frac{\min(\sqrt{v_R},\kappa_v)}{\kappa_v},
\frac{h_R}{H},
\frac{w_R}{W}
\right].
\label{eq:region_attributes}
\end{equation}
Here, $\kappa_a$ and $\kappa_v$ are the area-normalization and
variance clipping-and-normalization hyperparameters, respectively.
With the anchor-to-region difference
$d_R=\operatorname{LN}(\mu_R-a_R)$, the regional Carrier is defined as
\begin{equation}
c_R=
\begin{cases}
a_R, & |R|=1,\\[1mm]
a_R+\sigma(g)W_o
\operatorname{SiLU}\!\left(
\operatorname{LN}\!\left(
W_d d_R+\phi_\theta(s_R)
\right)\right), & |R|>1.
\end{cases}
\label{eq:carrier}
\end{equation}

Here, $W_d$ and $W_o$ are learnable linear projections,
$\phi_\theta$ is a learnable MLP embedding of $s_R$,
$g$ is a learnable scalar gate, and $\sigma$ is the sigmoid function.

The same Carrier parameters are shared across all regions, hierarchy levels,
and source views within a backbone.
Every Carrier is computed directly from the frozen source tokens within its
region rather than recursively from child Carriers.

\paragraph{Progressive hierarchy and frontier readout.}
TACC organizes each per-view source representation into a binary spatial
hierarchy $T_i$.
Starting from the complete spatial domain $\Omega_i$, each rectangular region
is bisected at the midpoint of its longer spatial dimension until each leaf
contains a single token.
To determine the refinement order, we measure the within-region scatter as
$D(R)=\sum_{p\in R}\|x_{i,p}-\mu_R\|_2^2$ and use
$\Delta(R)=D(R)-D(R_L)-D(R_R)$ as the marginal refinement gain of splitting
$R$ into $R_L$ and $R_R$.
Regions are refined in decreasing order of area and, among equal-area
regions, decreasing $\Delta(R)$, with exact ties resolved by stable ordering.

Starting from $\mathcal{F}_i(1)=\{\Omega_i\}$, each refinement replaces one
active region with its two children, increasing the frontier size by exactly
one while preserving complete spatial coverage.
Hence, every integer quota $q\in\{1,\ldots,P\}$ corresponds to a valid
frontier.
Let $H_i=(T_i,\{c_{i,R}\},\pi_i)$ denote the resulting hierarchy and its
refinement order.
For $q\in\{1,\ldots,P\}$, TACC reads the Carriers on the corresponding
active frontier as $Z_i(q)=\operatorname{Read}(H_i,q)$.
We additionally define
$Z_i(0)=\operatorname{Read}(H_i,0)=\varnothing$,
so that a zero quota simply omits source view $i$ from the target-time
readout.
At the full per-view quota $q=P$, the frontier contains only single-token
leaves and therefore exactly recovers the original source representation,
i.e., $Z_i(P)=X_i$.

\paragraph{Carrier training.}
Only the Carrier module is optimized, while all pretrained backbone
parameters remain frozen.
We use teacher--student distillation~\citep{hinton2015distilling} with a
view-subsampled but spatially dense teacher context $X^T\subseteq X$ to
obtain the teacher prediction
$I_t^T=\mathcal{D}(X^T,R_t)$, while the compressed representation $Z_t$
produces the student prediction
$I_t^S=\mathcal{D}(Z_t,R_t)$.

The training objective is
\begin{equation}
\mathcal{L}
=
\mathcal{L}_{\mathrm{render}}
+
\lambda_{\mathrm{dist}}\mathcal{L}_{\mathrm{distill}}
+
\lambda_{\mathrm{carrier}}\mathcal{L}_{\mathrm{carrier}}.
\label{eq:carrier_training}
\end{equation}

Here, $\mathcal{L}_{\mathrm{render}}$ combines $\ell_1$, MSE, and
image-gradient reconstruction losses;
$\mathcal{L}_{\mathrm{distill}}$ performs quality-gated RGB distillation
when the teacher is more accurate under the RGB reconstruction criterion;
and $\mathcal{L}_{\mathrm{carrier}}$ regularizes active non-leaf Carriers
toward their regional feature means.
Detailed Carrier architecture, training objectives, teacher construction,
backbone-specific protocols, and optimization settings are provided in
Appendix~\ref{app:carrier_training}.
\subsection{Target-Conditioned Evidence Allocation}
\label{sec:evidence_allocation}

Given a target view $t$, TACC distributes the fixed decoder-visible budget
$B$ across the source hierarchies through target-dependent per-view quotas
$\{q_i^t\}_{i=1}^{V}$.
The allocation stage is training-free and introduces no learnable parameters.

\paragraph{Target relevance and full-view preservation.}
TACC ranks source views by their geometric relevance to the target.
Let $d_i$ and $d_t$ denote the center-pixel ray directions, and
$\xi_i$ and $\xi_t$ the backbone-specific geometric descriptors used for
distance.
We define
\begin{equation}
r_i^t
=
\left\langle
\frac{d_i}{\|d_i\|_2},
\frac{d_t}{\|d_t\|_2}
\right\rangle
-
\lambda_d\,\delta(\xi_i,\xi_t).
\label{eq:target_relevance}
\end{equation}

For LagerNVS, LVSM, and CLiFT, we instantiate $\xi$ as the central-ray
Pl{\"u}cker moment, the central-ray origin, and the camera center from the
co-normalized c2w pose, respectively.
We compute
$\tilde{\delta}_i^t=\|\xi_i-\xi_t\|_2$
and min--max normalize it over the source views, with $\lambda_d=0.1$.
The highest-scoring source view
$i_t^\star=\arg\max_i r_i^t$
receives the full quota $P$.
The remaining views form the support set
$\mathcal{S}_t=\{1,\ldots,V\}\setminus\{i_t^\star\}$,
with remaining budget
$B_{\mathrm{rem}}=B-P$.

\paragraph{Entropy-adaptive evidence breadth.}

For the support views, let $\mu_{S_t}$ and $\sigma_{S_t}$ denote
the mean and population standard deviation of their relevance
scores. The standardized scores are
\begin{equation}
z_i^t
=
\begin{cases}
0,
& \sigma_{\mathcal{S}_t}\leq\epsilon_s,\\[1mm]
\dfrac{r_i^t-\mu_{\mathcal{S}_t}}
{\sigma_{\mathcal{S}_t}},
& \text{otherwise}.
\end{cases}
\label{eq:relevance_standardization}
\end{equation}

Here, $\epsilon_s$ is a small numerical threshold.
We then obtain
$p_i^t=\operatorname{Softmax}(z^t)_i$
and use the normalized entropy of this distribution to determine a shared
coverage floor:
\begin{equation}
H^t
=
-
\frac{
\sum_{i\in\mathcal{S}_t}
p_i^t\log p_i^t
}{
\log\!\bigl(\max(|\mathcal{S}_t|,2)\bigr)
},
\qquad
q_{\mathrm{floor}}^t
=
\left\lfloor
\frac{
H^t B_{\mathrm{rem}}
}{
|\mathcal{S}_t|
}
\right\rfloor .
\label{eq:entropy_floor}
\end{equation}

Higher entropy produces a larger shared floor and broader cross-view coverage,
whereas lower entropy leaves more of the budget for selective allocation.

\paragraph{Relevance-weighted residual allocation.}
After assigning the shared floor, the residual budget is
$B_{\mathrm{res}}^t
=
B_{\mathrm{rem}}
-
|\mathcal{S}_t|q_{\mathrm{floor}}^t$.
We apply a softmax with temperature $\tau=0.8$ over the support set to obtain
$\hat{p}_i^t$, yielding soft residual quotas
$a_i^t=B_{\mathrm{res}}^t\hat{p}_i^t$.
We then use capacity-aware largest-remainder rounding to convert the soft
residual quotas into integer allocations $b_i^t$ under the remaining
per-view capacity $P-q_{\mathrm{floor}}^t$, while preserving the residual
budget exactly.

The final quotas are
$q_{i_t^\star}^t=P$ and
$q_i^t=q_{\mathrm{floor}}^t+b_i^t$
for $i\in\mathcal{S}_t$.
TACC then reads
$Z_i^t=\operatorname{Read}(H_i,q_i^t)$
from each source hierarchy and concatenates the resulting representations to
form $Z_t$, which is passed to the unchanged target-conditioned rendering
pathway of the corresponding backbone.

%% file: sections/4_experiments.tex
\section{Experiments}
\label{sec:experiments}

\subsection{Experimental Settings}
\label{sec:experimental_settings}

\paragraph{Datasets.}
We evaluate TACC on DL3DV~\citep{dl3dv} and
RealEstate10K (RE10K)~\citep{realestate10k}.
On DL3DV, the Carrier modules are trained on 10,139 training scenes and
evaluated on the official 140-scene benchmark.
On RE10K, the LVSM Carrier is trained using scenes from the RE10K training split.
For the large-view RE10K evaluation, we select scenes with more than
200 available views from the 6,474 scenes in our preprocessed test split,
yielding 1,633 evaluation scenes that are disjoint from those used for
training and validation.

\paragraph{Backbones, baselines, and metrics.}
We integrate TACC into three latent NVS backbones:
LagerNVS~\citep{lagernvs}, LVSM Decoder-Only~\citep{lvsm},
and CLiFT~\citep{clift}.
We compare against ToMe~\citep{tome}, DivPrune~\citep{divprune},
and ZPressor~\citep{zpressor}.
Within each experimental setting, all compression methods operate under
the same decoder-visible token budget and use identical scene IDs and
source- and target-view indices.
Full Context denotes inference with all available source tokens using
the corresponding frozen pretrained backbone, without dense-context
adaptation or fine-tuning.
For CLiFT, Native denotes its original compression pathway retrained
under matched experimental settings, with the pretrained encoder and
renderer kept frozen.
We evaluate reconstruction quality using PSNR, SSIM~\citep{ssim},
and LPIPS~\citep{lpips}, and measure target-time efficiency using
Effective FPS.
For a fixed set of source views, Effective FPS includes all
target-dependent computation from the arrival of the target camera to
the final RGB prediction, but excludes one-time source encoding and,
for TACC, one-time Carrier construction.
Amortized end-to-end (E2E) FPS includes these one-time costs;
see Appendix~\ref{app:runtime}.

\paragraph{Implementation details.}
We use the official implementations and pretrained checkpoints of
LagerNVS, LVSM Decoder-Only, and CLiFT, and train only the Carrier module
while keeping the pretrained backbone parameters frozen.
For CLiFT, TACC replaces its native compression pathway rather than
introducing an additional compression stage.
LagerNVS and LVSM Decoder-Only are evaluated at $256\times256$, while
CLiFT follows its native $256\times448$ resolution.
For each backbone, a single Carrier checkpoint is shared across all
evaluated source-view counts and target views under its corresponding
decoder-visible budget.
Backbone-specific training protocols, optimization settings, and
checkpoint-selection criteria are provided in
Appendix~\ref{app:carrier_training}.
All experiments are conducted on NVIDIA A800 GPUs.

\subsection{Matched-Budget Comparisons and Efficiency Analysis}
\label{sec:main_comparisons}

\paragraph{Matched-budget comparisons.}
We compare TACC with representation-reduction baselines on DL3DV using
LagerNVS with source-view counts ranging from 80 to 120.
Compressed methods use a fixed $A=10$ decoder-visible budget, while
Full Context uses all source tokens; the target-view batch size is fixed
to 4 for all methods.

\begin{figure}[t]
    \vspace{-8pt}
    \centering
    \includegraphics[width=\linewidth]{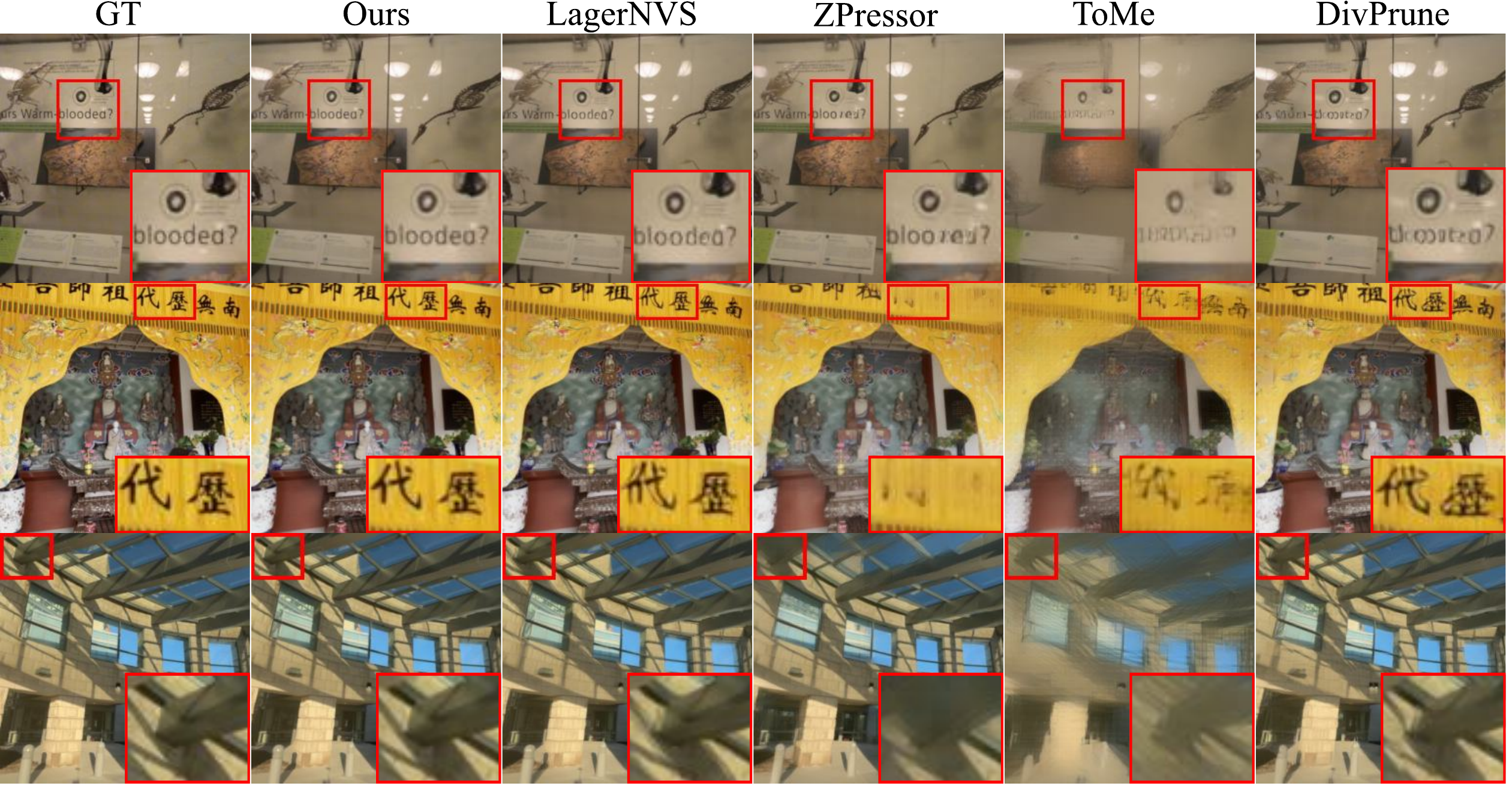}
    \vspace{-20pt}
    \caption{
        \textbf{Qualitative comparison on DL3DV with 100 source views.}
        Compressed methods use a fixed $A=10$ decoder-visible budget,
        while LagerNVS uses Full Context.
    }
    \label{fig:qualitative_dl3dv}
    \vspace{-10pt}
\end{figure}

As shown in Tab.~\ref{tab:main_comparison}, TACC achieves the strongest
reconstruction quality among compressed methods while maintaining nearly
constant target-time throughput as the source set grows. Although ToMe and
DivPrune are slightly faster, they incur substantially larger quality
degradation. Notably, TACC also yields higher reconstruction quality than
the frozen Full Context checkpoint in all three evaluated dense-view
settings. For the frozen pretrained LagerNVS checkpoint evaluated here,
Full Context quality decreases as the test-time source context grows,
whereas TACC operates under a fixed decoder-visible budget and maintains
higher reconstruction quality.
Fig.~\ref{fig:qualitative_dl3dv} provides consistent qualitative evidence.
These comparisons serve different purposes: Full Context
provides a reference for inference with the released frozen backbone,
whereas matched-budget comparisons among compression methods more
directly assess compression effectiveness.

\begin{table}[t]
    \centering
    \vspace{-10pt}
    \caption{
        \textbf{Quantitative comparisons on DL3DV.}
        Compression methods use a fixed $A=10$ decoder-visible budget
        and identical source views within each setting;
        \colorbox{gray!18}{Full Context} uses all LagerNVS source tokens
        as the uncompressed reference.
    }
    
    \label{tab:main_comparison}

    \small
    \setlength{\tabcolsep}{3.0pt}
    \renewcommand{\arraystretch}{1.08}

    \resizebox{\linewidth}{!}{
    \begin{tabular}{@{}lcccccccccccc@{}}

        \toprule

        &
        \multicolumn{4}{c}{\textbf{80 Source Views}}
        &
        \multicolumn{4}{c}{\textbf{100 Source Views}}
        &
        \multicolumn{4}{c}{\textbf{120 Source Views}}
        \\

        \cmidrule(lr){2-5}
        \cmidrule(lr){6-9}
        \cmidrule(lr){10-13}

        \textbf{Method}
        & PSNR $\uparrow$
        & SSIM $\uparrow$
        & LPIPS $\downarrow$
        & Eff. FPS $\uparrow$

        & PSNR $\uparrow$
        & SSIM $\uparrow$
        & LPIPS $\downarrow$
        & Eff. FPS $\uparrow$

        & PSNR $\uparrow$
        & SSIM $\uparrow$
        & LPIPS $\downarrow$
        & Eff. FPS $\uparrow$
        \\

        \midrule
        \multicolumn{13}{c}{\textbf{LagerNVS-based}} \\
        \midrule

        \rowcolor{gray!18}
        Full Context
        & 25.6766
        & 0.8064
        & 0.0963
        & 4.40

        & 25.2932
        & 0.7953
        & 0.1012
        & 3.57

        & 24.9604
        & 0.7851
        & 0.1056
        & 2.99
        \\
        \hdashline

        ToMe
        & 20.0344
        & 0.5442
        & 0.3548
        & 31.43

        & 19.7096
        & 0.5304
        & 0.3737
        & 31.00

        & 19.2548
        & 0.5122
        & 0.3991
        & 31.34
        \\

        DivPrune
        & 21.7624
        & 0.6560
        & 0.2053
        & \textbf{31.54}

        & 21.5944
        & 0.6479
        & 0.2106
        & 31.07

        & 21.4627
        & 0.6420
        & 0.2147
        & 31.08
        \\

        ZPressor
        & 24.1568
        & 0.7720
        & 0.1381
        & 31.26

        & 24.0773
        & 0.7684
        & 0.1396
        & \textbf{31.61}

        & 24.5947
        & 0.7872
        & 0.1338
        & \textbf{31.52}
        \\

        \textbf{Ours}
        & \textbf{27.3415}
        & \textbf{0.8649}
        & \textbf{0.0791}
        & 27.95

        & \textbf{26.8787}
        & \textbf{0.8523}
        & \textbf{0.0833}
        & 28.12

        & \textbf{26.4732}
        & \textbf{0.8416}
        & \textbf{0.0869}
        & 27.96
        \\

        \bottomrule

    \end{tabular}
    }
\end{table}

\paragraph{Cross-dataset generalization.}
We directly evaluate the DL3DV-trained LagerNVS and TACC checkpoints on
RE10K without fine-tuning, using 80 to 120 source views, a fixed
$A=10$ decoder-visible budget, and 12 target views.
Across these settings, TACC consistently outperforms matched-budget
compression baselines in reconstruction quality while maintaining nearly
constant Effective FPS, whereas frozen Full Context inference becomes
progressively slower.

\begin{figure}[H]
    
    \centering
    \includegraphics[width=\linewidth]{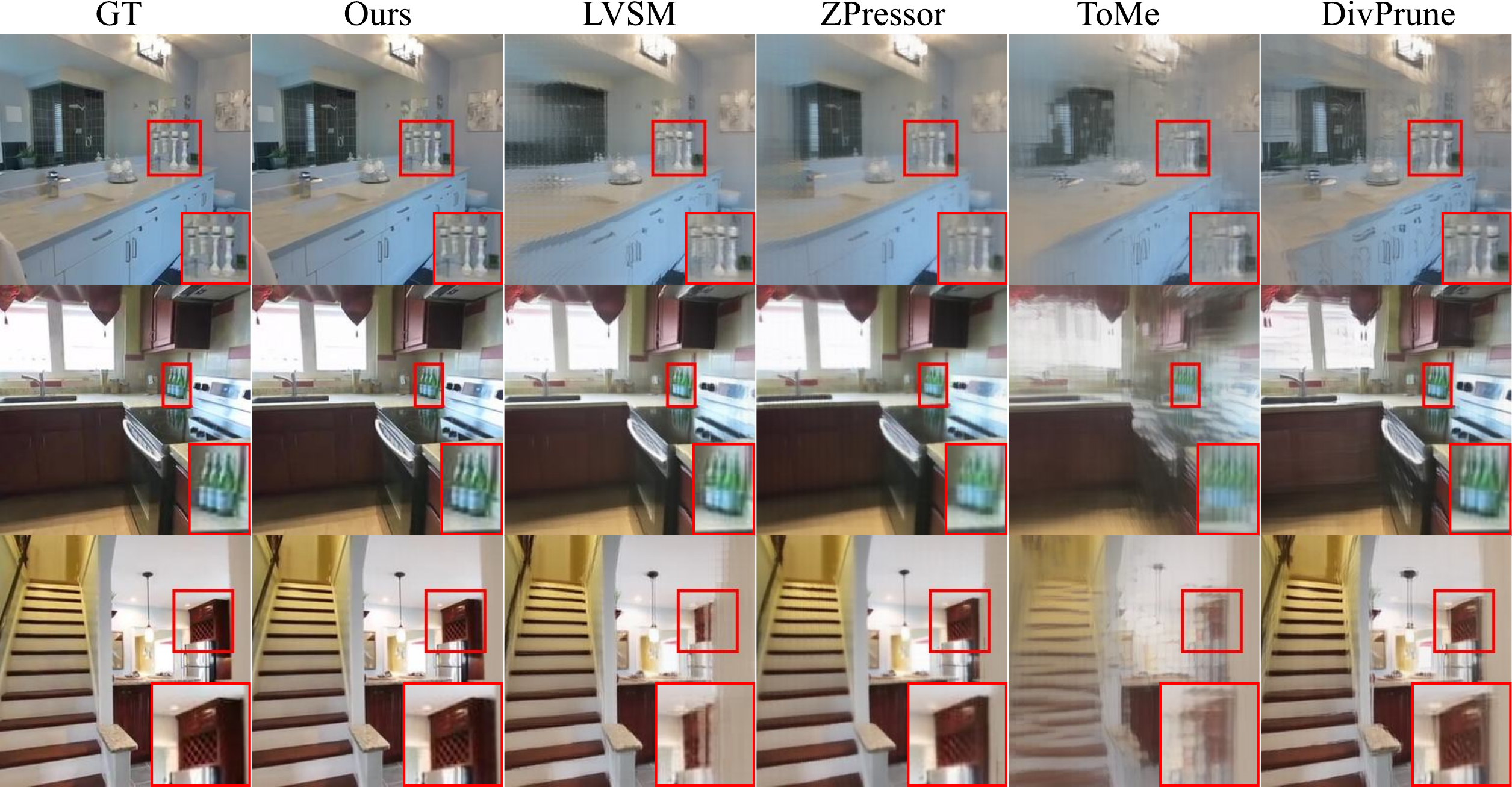}
    \vspace{-5pt}
    \caption{
        \textbf{Qualitative comparison with LVSM Decoder-Only on RE10K.}
        Compressed methods use a fixed $A=10$ decoder-visible budget,
        while LVSM Decoder-Only uses Full Context.
    }
    \label{fig:qualitative_lvsm}
    
\end{figure}
\begin{figure}[H]
    \centering
    \vspace{-5pt}
    \includegraphics[
        width=\linewidth
    ]{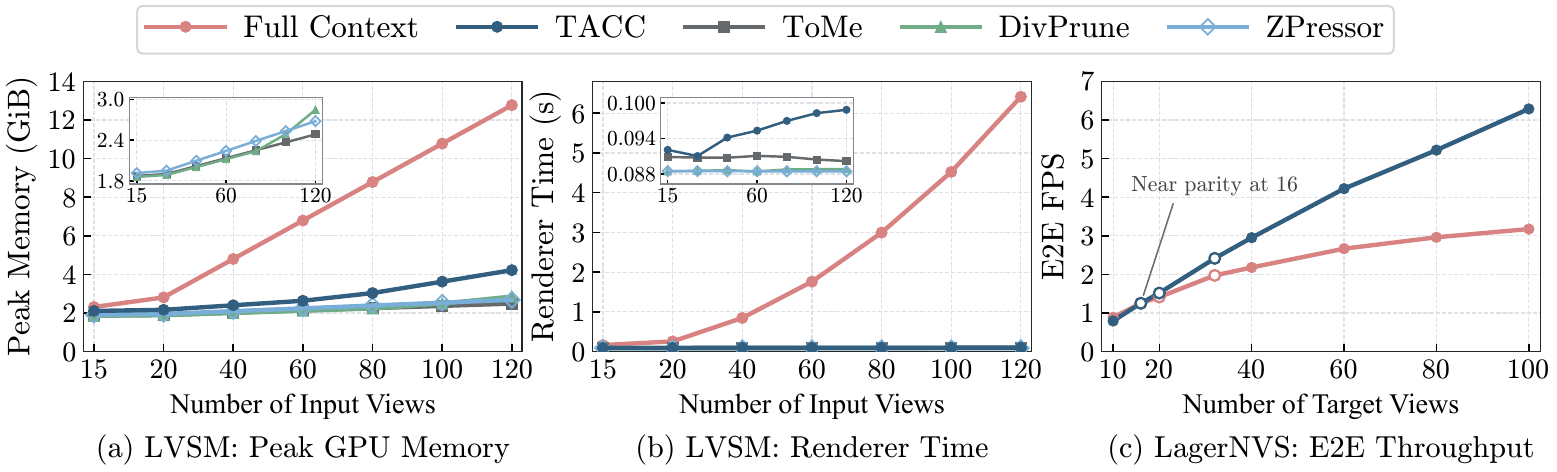}
    
    \caption{
        \textbf{Efficiency analysis.}
        We report peak GPU memory and renderer latency for LVSM on RE10K
        with source-view counts ranging from 15 to 120, and amortized E2E FPS
        for LagerNVS on DL3DV with 80 source views across increasing target counts.
    }
    \label{fig:lvsm_efficiency}
    
    \vspace{-8pt}
\end{figure}

\paragraph{Cross-backbone evaluation and efficiency.}
We evaluate TACC with LVSM Decoder-Only under the same RE10K
protocol. Only the Carrier module is trained, while the pretrained
backbone remains frozen.
As shown in Tab.~\ref{tab:re10k_scaling}, with source-view counts ranging
from 80 to 120, TACC achieves the highest reconstruction quality among
compressed methods and improves both reconstruction quality and target-time
throughput over frozen Full Context, while maintaining Effective FPS near 10.
We further compare TACC with a dense-context-fine-tuned
LVSM in Appendix~\ref{app:lvsm_dense_scaling}, where TACC maintains
slightly higher reconstruction quality and substantially greater
target-time rendering throughput without backbone fine-tuning.
Fig.~\ref{fig:qualitative_lvsm} further shows that TACC better preserves
fine structures under the matched budget.
For LVSM Decoder-Only, Figs.~\ref{fig:lvsm_efficiency}(a)
and~\ref{fig:lvsm_efficiency}(b) show that Full Context incurs increasing
renderer latency and peak GPU memory as source views grow, whereas
TACC maintains nearly constant renderer latency and slower memory growth.
At 120 source views, TACC reduces peak allocated memory from
12.77 to 4.22~GiB, a 66.96\% reduction.

For a fixed source set, source encoding and Carrier construction are
performed once and reused across target views. Effective FPS therefore
measures sustained target-time throughput, whereas amortized end-to-end
(E2E) FPS in Fig.~\ref{fig:lvsm_efficiency}(c) also includes these
one-time costs. With 80 source views, TACC is slightly slower than
Full Context at 10 target views and reaches approximately equal E2E
throughput at 16 targets. At 20 targets, TACC achieves approximately
7.3\% higher E2E FPS, with its advantage reaching $1.98\times$ at
100 targets. For LagerNVS on DL3DV, Tab.~\ref{tab:main_runtime_costs}
reports target-time latency, Carrier construction time, and reusable
source-cache sizes. Across 80--120 source views, TACC maintains
approximately 32~ms decoder latency, compared with 227--334~ms for
Full Context, while incurring additional construction and cache costs.
Additional LVSM results with larger source-view counts and runtime details
are provided in Appendices~\ref{app:lvsm_dense_scaling}
and~\ref{app:runtime}.

\begin{table}[H]
    \vspace{-10pt}
    \centering
    \caption{
        \textbf{Quantitative comparisons on RE10K.}
        Results are reported for LagerNVS and LVSM Decoder-Only, with compressed
        methods using a fixed $A=10$ decoder-visible budget and
        \colorbox{gray!18}{Full Context} using all source tokens as the
        uncompressed reference.
    }
    \label{tab:re10k_scaling}

    \small
    \setlength{\tabcolsep}{2.5pt}
    \renewcommand{\arraystretch}{1.08}

    \resizebox{\linewidth}{!}{
    \begin{tabular}{@{}lcccccccccccc@{}}

        \toprule

        &
        \multicolumn{4}{c}{\textbf{80 Source Views}}
        &
        \multicolumn{4}{c}{\textbf{100 Source Views}}
        &
        \multicolumn{4}{c}{\textbf{120 Source Views}}
        \\

        \cmidrule(lr){2-5}
        \cmidrule(lr){6-9}
        \cmidrule(lr){10-13}

        \textbf{Method}
        & PSNR $\uparrow$
        & SSIM $\uparrow$
        & LPIPS $\downarrow$
        & Eff. FPS $\uparrow$

        & PSNR $\uparrow$
        & SSIM $\uparrow$
        & LPIPS $\downarrow$
        & Eff. FPS $\uparrow$

        & PSNR $\uparrow$
        & SSIM $\uparrow$
        & LPIPS $\downarrow$
        & Eff. FPS $\uparrow$
        \\

        \midrule
        \multicolumn{13}{c}{\textbf{LagerNVS-based}} \\
        \midrule

        \rowcolor{gray!18}
        Full Context
        & 28.5991
        & 0.9066
        & 0.0681
        & 4.38

        & 28.0454
        & 0.8976
        & 0.0718
        & 3.54

        & 27.5702
        & 0.8890
        & 0.0754
        & 2.97
        \\
        \hdashline

        ToMe
        & 22.1035
        & 0.7030
        & 0.2889
        & \textbf{31.51}

        & 21.8357
        & 0.6915
        & 0.3070
        & 31.05

        & 21.4481
        & 0.6780
        & 0.3276
        & \textbf{31.50}
        \\

        DivPrune
        & 25.1277
        & 0.8298
        & 0.1215
        & 31.46

        & 24.9600
        & 0.8252
        & 0.1240
        & \textbf{31.42}

        & 24.8335
        & 0.8217
        & 0.1259
        & 31.12
        \\

        ZPressor
        & 27.9540
        & 0.8999
        & 0.0854
        & 31.28

        & 27.7685
        & 0.8961
        & 0.0868
        & 31.41

        & 27.5760
        & 0.8921
        & 0.0882
        & 31.20
        \\

        \textbf{Ours}
        & \textbf{30.2044}
        & \textbf{0.9285}
        & \textbf{0.0645}
        & 27.47

        & \textbf{29.4752}
        & \textbf{0.9182}
        & \textbf{0.0681}
        & 27.42

        & \textbf{28.8774}
        & \textbf{0.9090}
        & \textbf{0.0715}
        & 27.44
        \\

        \midrule
        \multicolumn{13}{c}{\textbf{LVSM Decoder-Only-based}} \\
        \midrule

        \rowcolor{gray!18}
        Full Context
        & 22.7867
        & 0.7489
        & 0.2134
        & 0.33

        & 22.4844
        & 0.7364
        & 0.2267
        & 0.22

        & 22.2624
        & 0.7269
        & 0.2375
        & 0.16
        \\
        \hdashline

        ToMe
        & 19.4181
        & 0.5842
        & 0.3874
        & \textbf{11.25}

        & 19.1861
        & 0.5698
        & 0.4015
        & 11.12

        & 18.8588
        & 0.5585
        & 0.4182
        & \textbf{11.34}
        \\

        DivPrune
        & 20.3286
        & 0.6110
        & 0.3084
        & 11.21

        & 20.0773
        & 0.5990
        & 0.3233
        & \textbf{11.29}

        & 19.9148
        & 0.5911
        & 0.3338
        & 11.10
        \\

        ZPressor
        & 27.8870
        & 0.8883
        & 0.1047
        & 11.17

        & 27.8993
        & 0.8885
        & 0.1045
        & 11.28

        & 27.9235
        & 0.8890
        & 0.1041
        & 11.25
        \\

        \textbf{Ours}
        & \textbf{30.2851}
        & \textbf{0.9260}
        & \textbf{0.0739}
        & 10.28

        & \textbf{30.1978}
        & \textbf{0.9250}
        & \textbf{0.0744}
        & 10.15

        & \textbf{30.1441}
        & \textbf{0.9243}
        & \textbf{0.0750}
        & 10.03
        \\

        \bottomrule

    \end{tabular}
    }
    \vspace{-10pt}
    
\end{table}

\begin{figure}[H]
    \centering
    \includegraphics[
        width=\linewidth
    ]{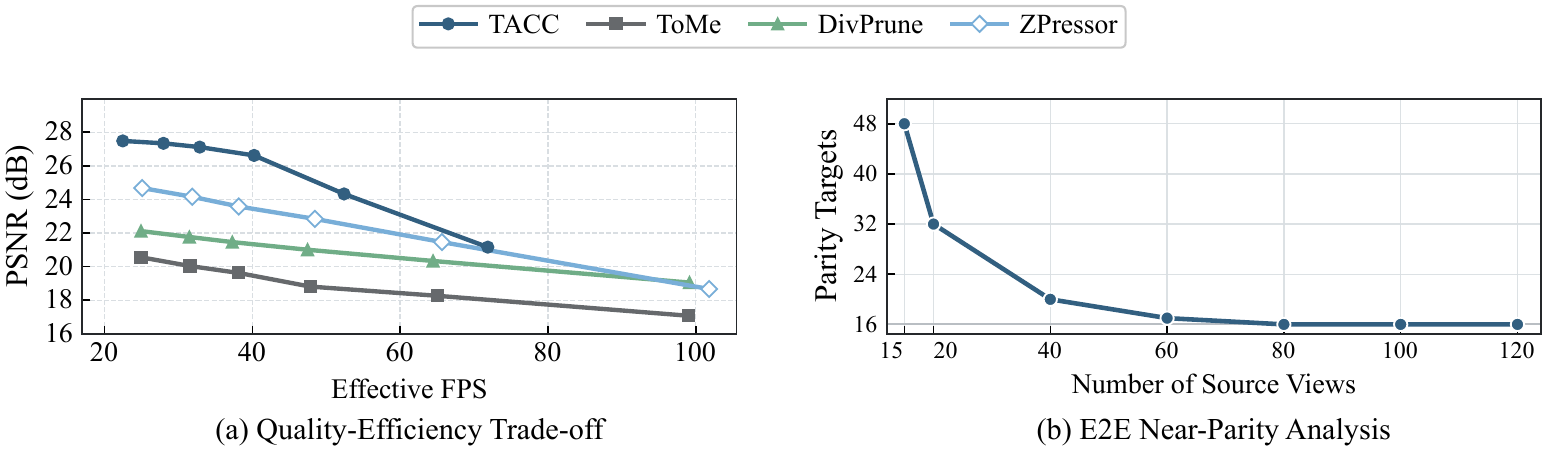}
    \vspace{-20pt}
    
    \caption{
        \textbf{Quality--efficiency and E2E near-parity analysis.}
        We report PSNR versus Effective FPS across six budgets
        with 80 source views (a), and measured E2E near-parity
        target counts across 15--120 source views at $A=10$ (b),
        using LagerNVS on DL3DV.
    }
    \label{fig:tacc_quality_efficiency}
    
\end{figure}

\paragraph{Quality--efficiency and E2E amortization.}
We further examine LagerNVS on DL3DV across six
matched decoder-visible budgets. As shown in
Fig.~\ref{fig:tacc_quality_efficiency}(a), TACC achieves
higher PSNR than the competing compression methods at
every evaluated budget. Notably, TACC at $A=6$ achieves
26.62~dB and 39.52 Effective FPS, outperforming ZPressor
at $A=13$ (24.68~dB and 25.09 Effective FPS) in both metrics.
Fig.~\ref{fig:tacc_quality_efficiency}(b) additionally
shows that the measured near-parity target count decreases
from 48 at 15 source views to 16 at 80--120 source views.
These are approximate parity reference points rather than
exact E2E break-even thresholds. Detailed protocols and
quantitative results are provided in
Appendices~\ref{app:tighter_budget} and~\ref{app:runtime}.%

\begin{table}[H]
    \vspace{-15pt}
    \centering
    \caption{
        \textbf{Target-time runtime and one-time costs on DL3DV.}
        Results use LagerNVS with a fixed $A=10$ budget.
        Allocation/readout and decoder latencies are in milliseconds
        per target view; Carrier construction is in seconds per scene,
        and reusable source-cache sizes are in GiB.
    }
    \label{tab:main_runtime_costs}

    \footnotesize
    \setlength{\tabcolsep}{3.5pt}
    \renewcommand{\arraystretch}{1.08}
    \begin{tabular}{@{}ccccccc@{}}
        \toprule
        \textbf{Views}
        & \shortstack{\textbf{Alloc.+Read.}\\\textbf{(ms)}}
        & \shortstack{\textbf{TACC Dec.}\\\textbf{(ms)}}
        & \shortstack{\textbf{Full Dec.}\\\textbf{(ms)}}
        & \shortstack{\textbf{Build}\\\textbf{(s)}}
        & \shortstack{\textbf{TACC Cache}\\\textbf{(GiB)}}
        & \shortstack{\textbf{Full Cache}\\\textbf{(GiB)}} \\
        \midrule
        80  & 3.045 & 32.737 & 227.035 & 3.1323 & 1.6859 & 0.3133 \\
        100 & 3.094 & 32.468 & 279.945 & 3.9474 & 2.1074 & 0.3917 \\
        120 & 3.344 & 32.419 & 334.353 & 4.7076 & 2.5288 & 0.4700 \\
        \bottomrule
    \end{tabular}
\end{table}

\vspace{-5pt}

\vspace{-20pt}
\begin{table}[H]
    \vspace{-3pt}
    \centering
    \caption{
        \textbf{Comparison on CLiFT on DL3DV with increasing source-view counts.}
        Compressed variants use a fixed $A=6$ budget, while
        \colorbox{gray!18}{Full Context} uses the complete representation.
    }
    \label{tab:clift_scaling}

    \small
    \setlength{\tabcolsep}{2.5pt}
    \renewcommand{\arraystretch}{1.08}

    \resizebox{\linewidth}{!}{
    \begin{tabular}{@{}lcccccccccccc@{}}

        \toprule

        &
        \multicolumn{4}{c}{\textbf{24 Source Views}}
        &
        \multicolumn{4}{c}{\textbf{36 Source Views}}
        &
        \multicolumn{4}{c}{\textbf{48 Source Views}}
        \\

        \cmidrule(lr){2-5}
        \cmidrule(lr){6-9}
        \cmidrule(lr){10-13}

        \textbf{Method}
        & PSNR $\uparrow$
        & SSIM $\uparrow$
        & LPIPS $\downarrow$
        & Eff. FPS $\uparrow$

        & PSNR $\uparrow$
        & SSIM $\uparrow$
        & LPIPS $\downarrow$
        & Eff. FPS $\uparrow$

        & PSNR $\uparrow$
        & SSIM $\uparrow$
        & LPIPS $\downarrow$
        & Eff. FPS $\uparrow$
        \\

        \midrule
        \multicolumn{13}{c}{\textbf{CLiFT-based}} \\
        \midrule

        \rowcolor{gray!18}
        Full Context
        & 20.1825
        & 0.5962
        & 0.2912
        & 15.43

        & 20.1843
        & 0.5940
        & 0.2992
        & 10.64

        & 20.1703
        & 0.5930
        & 0.3014
        & 8.08
        \\
        \hdashline

        Native
        & 21.8549
        & 0.6574
        & 0.2569
        & 47.29

        & 21.7869
        & 0.6513
        & 0.2630
        & \textbf{47.46}

        & 21.7493
        & 0.6498
        & 0.2647
        & 47.39
        \\

        ToMe
        & 19.4351
        & 0.5367
        & 0.3850
        & 47.27

        & 19.3258
        & 0.5292
        & 0.3963
        & 47.19

        & 18.7800
        & 0.5031
        & 0.4332
        & \textbf{47.40}
        \\

        DivPrune
        & 19.4529
        & 0.5515
        & 0.3264
        & 47.21

        & 19.3823
        & 0.5437
        & 0.3371
        & 47.13

        & 19.3468
        & 0.5411
        & 0.3407
        & 47.09
        \\

        ZPressor
        & 20.9791
        & 0.6344
        & 0.2814
        & \textbf{47.43}

        & 21.0250
        & 0.6348
        & 0.2825
        & 47.43

        & 20.9394
        & 0.6307
        & 0.2855
        & 47.27
        \\

        \textbf{Ours}
        & \textbf{23.3425}
        & \textbf{0.7397}
        & \textbf{0.2180}
        & 46.37

        & \textbf{23.5100}
        & \textbf{0.7441}
        & \textbf{0.2165}
        & 45.20

        & \textbf{23.5842}
        & \textbf{0.7470}
        & \textbf{0.2149}
        & 43.18
        \\

        \bottomrule

    \end{tabular}
    }

    \vspace{-15pt}
\end{table}

\paragraph{Complementary evaluation on CLiFT.}
We further evaluate TACC with CLiFT~\citep{clift} on DL3DV.
TACC replaces its native compression-and-retrieval pathway with
progressive per-view Carrier encoding and target-conditioned frontier
readout, while retaining the pretrained encoder and renderer.
As shown in Tab.~\ref{tab:clift_scaling}, TACC achieves the best
reconstruction quality among compressed variants across 24, 36, and 48
source views, outperforming the retrained Native pathway by
1.49--1.83\,dB in PSNR with consistent SSIM and LPIPS improvements.
The quality margin increases as the source-to-budget ratio grows from
$4\times$ to $8\times$, with only a modest reduction in target-time
throughput.
Appendix~\ref{app:clift_qualitative} provides qualitative comparisons
in Fig.~\ref{fig:app_clift_qualitative} and reports the official
pretrained Native checkpoint results in Tab.~\ref{tab:clift_official}.

\subsection{Ablation Study}
\label{sec:ablation}

We evaluate TACC using LagerNVS on DL3DV with 80 source views and a fixed
$A=10$ decoder-visible budget. Tab.~\ref{tab:component_ablation} shows
that all controlled variants reduce reconstruction quality relative to TACC,
while the qualitative comparisons in Fig.~\ref{fig:qual_ablation} exhibit
consistent trends. Together, these results support the contributions of both
learned regional Carriers and target-conditioned evidence allocation, with
further controls provided in Appendix~\ref{app:allocation_ablations}.

\vspace{-10pt}

\noindent
\begin{minipage}[t]{0.49\linewidth}
    \vspace{0pt}
    \setlength{\parskip}{0pt}

    The \emph{w/o Carrier} variant retains TACC's target-conditioned
    per-view quotas but replaces Carrier readout with approximately
    uniform spatial subsampling of the original tokens, while
    \emph{Uniform Alloc.} instead uses approximately uniform per-view
    quotas, with its Carrier retrained under the same allocation rule.
    \emph{Top-$A$ Full Views} retains the $A$ highest-ranked complete
    source views according to TACC's relevance score.

\end{minipage}
\hfill
\begin{minipage}[t]{0.49\linewidth}
    \vspace{-10pt}
    \centering

    \captionof{table}{
        \textbf{Component ablations.}
    }
    \label{tab:component_ablation}

    \vspace{1pt}

    \small
    \setlength{\tabcolsep}{3.0pt}
    \renewcommand{\arraystretch}{1.04}

    \resizebox{\linewidth}{!}{
        \begin{tabular}{@{}lccc@{}}
            \toprule
            Variant
            & PSNR $\uparrow$
            & SSIM $\uparrow$
            & LPIPS $\downarrow$
            \\
            \midrule

            Top-$A$ Full Views
            & 21.6315
            & 0.6509
            & 0.1582
            \\

            Anchor-only Carrier
            & 22.2410
            & 0.6936
            & 0.1316
            \\

            w/o Carrier
            & 25.1493
            & 0.8047
            & 0.0984
            \\

            Uniform Alloc.
            & 23.0085
            & 0.7154
            & 0.1945
            \\

            \textbf{TACC}
            & \textbf{27.3415}
            & \textbf{0.8649}
            & \textbf{0.0791}
            \\

            \bottomrule
        \end{tabular}
    }

\end{minipage}

\vspace{0pt}

\noindent
\begin{minipage}[t]{0.68\linewidth}
    \vspace{0pt}
    \centering

    \includegraphics[
        width=\linewidth
    ]{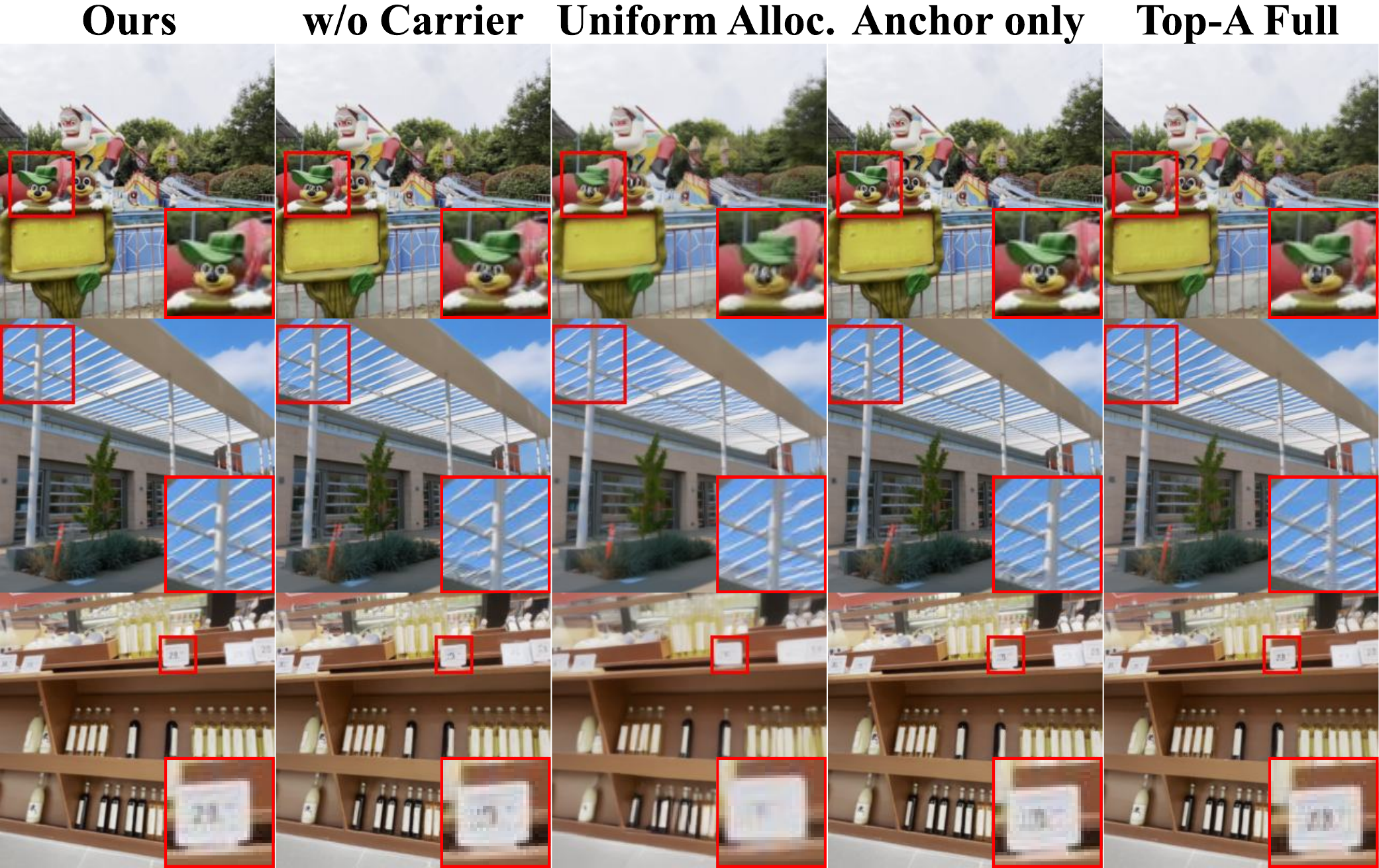}

    \vspace{-6pt}

    \captionof{figure}{
        \textbf{Qualitative analysis of Carrier and evidence allocation.}
        We compare TACC with w/o Carrier, Uniform Alloc.,
        Anchor-only Carrier, and Top-$A$ Full Views on DL3DV.
        Red boxes indicate the regions enlarged in the corresponding insets.
    }
    \label{fig:qual_ablation}

\end{minipage}
\hfill
\begin{minipage}[t]{0.30\linewidth}
    \vspace{0pt}
    \setlength{\parskip}{0pt}

    \justifying
    \hyphenpenalty=50
    \exhyphenpenalty=50
    \tolerance=1000
    \emergencystretch=0.5em

    \emph{Anchor-only Carrier} preserves the hierarchy and allocation
    but replaces learned regional Carriers with their anchors.
    Top-$A$ Full Views indicates that selecting target-relevant complete
    views alone is insufficient, while Anchor-only Carrier shows that
    the hierarchy and allocation without learned regional Carriers cannot
    recover TACC's performance. The degradation under
    \emph{w/o Carrier} further supports the contribution of learned
    regional Carriers, whereas the larger drop under
    \emph{Uniform Alloc.} highlights the role of target-conditioned
    evidence allocation.

\end{minipage}

\vspace{2pt}


\noindent\textbf{Additional component and loss ablations.}
We further examine two allocation safeguards and the training
objectives. With the Carrier checkpoint fixed, removing Top-1
full-view preservation or the entropy-based coverage floor
reduces PSNR by 4.20 and 2.57 dB, respectively.
Separately retraining the Carrier without the distillation
or Carrier loss leads to average PSNR drops of 0.56 and
0.62 dB, respectively, across 80--120 source views.
Detailed results are provided in
Appendix~\ref{app:allocation_ablations}.%

%% file: sections/5_conclusion.tex
\section{Conclusion}

We addressed the challenge of efficiently processing dense source views in
feed-forward latent NVS, where growing scene-token context incurs substantial
target-time memory and latency overhead. We introduced TACC, a target-aware
compression module that organizes each source view into a reusable progressive
Carrier hierarchy. Under a fixed decoder-visible token budget, TACC assigns
exact per-view quotas according to source-target geometric relevance and reads
the corresponding Carrier frontiers to construct a compact target-conditioned
representation while keeping the pretrained rendering pathway unchanged.
Across three latent NVS backbones and two datasets, TACC consistently improves
reconstruction quality over matched-budget compression baselines while
maintaining stable target-time throughput under dense source views,
demonstrating the effectiveness of target-aware representation-granularity
allocation under a fixed rendering budget.

%% file: sections/6_statements.tex
\section*{Reproducibility Statement}

We provide the main algorithmic details of TACC in Secs.~3.2 and~3.3,
including progressive Carrier construction, frontier readout, and
target-conditioned evidence allocation.
The datasets, pretrained backbones, decoder-visible budgets, image
resolutions, evaluation metrics, and hardware settings are described in
Sec.~4.1.

The appendix provides additional details required for reproduction,
including the training objectives, backbone-specific training protocols,
validation and checkpoint-selection criteria, optimization settings,
target-time inference procedure, backbone integration, deterministic
source/target view construction, and exact-budget baseline implementations.
It further reports runtime analysis and additional robustness and
source-view-count evaluations.

All experiments use publicly available datasets and official pretrained
backbones where applicable.
We plan to release the implementation and trained Carrier checkpoints to
facilitate reproduction of the reported results.

\section*{AI Use Statement}

We used GPT 5.6 to assist with manuscript writing and language refinement,
including translation, improving clarity and readability, and organizing the
presentation of the paper.
It was also used to provide feedback on the presentation and interpretation
of experimental results.
All technical content, experimental results, claims, and AI-assisted text
were reviewed and verified by the authors, who take full responsibility for
the final content of the paper.

%% file: sections/appendix.tex
\appendix

\section{Limitations and Future Work}
\label{app:limitations}

TACC introduces one-time computation and storage costs to construct
and cache the per-view Carrier hierarchies. Although these costs are
amortized across targets, they are excluded from Effective FPS, and
the end-to-end benefit depends on the number of targets rendered
from the same source context. The LagerNVS E2E results in Fig.~\ref{fig:lvsm_efficiency}(c)
and the complementary LVSM evaluation in
Appendix~\ref{app:runtime} illustrate this trade-off.
In the evaluated LagerNVS setting, TACC is slightly
slower than Full Context with 10 targets but achieves higher E2E
throughput as more targets are rendered. Since the one-time
overhead is less effectively amortized for shorter target sequences,
reducing hierarchy construction cost could improve E2E performance
in such workloads, while reducing the cache footprint would lower
storage requirements.

TACC currently uses target relevance to determine the token budget
for each source view, while its intra-view frontier follows a
precomputed refinement order. Thus, the selected regional
granularity does not explicitly adapt to the content most relevant
to the current target.

These spatial and geometric inductive biases may limit representation
flexibility under tighter token budgets, but enable reusable
source-side compression and efficient target-conditioned readout. A promising direction is to improve
target-conditioned regional refinement within each fixed per-view
quota, allowing the same number of tokens to represent
target-relevant information more effectively. Such an extension
could build on the reusable Carrier hierarchies while preserving
the exact decoder-visible budget and limiting additional
target-time overhead.

Investigating the effective limit of useful source-view information
across datasets and viewing configurations may also inspire
heuristic acceleration strategies that avoid processing
redundant observations.

\section{Additional Implementation Details}
\label{app:implementation}

\subsection{Carrier Architecture and Training Details}
\label{app:carrier_training}

\paragraph{Carrier architecture.}
All three backbones use the same Carrier architecture with token dimension
$C=768$ and hidden dimension $D_h=192$.
For the regional attributes in Eq.~(\ref{eq:region_attributes}), we set
$\kappa_a=8$ and $\kappa_v=10$.
The four-dimensional attribute embedding is
$\phi_\theta:
\operatorname{Linear}(4,32)\rightarrow
\operatorname{SiLU}\rightarrow
\operatorname{Linear}(32,192)$.

The experimental checkpoints retain a duplicate of the first area
attribute: for $s_R=[s_1,s_2,s_3,s_4]$, their input is
$\tilde{s}_R=[s_1,s_1,s_2,s_3,s_4]$.
This redundant input is exactly equivalent to the four-dimensional
formulation. Specifically, the first-layer weights
$\widetilde{W}_{\phi}^{(1)}
=[w_1,w_2,w_3,w_4,w_5]$
can be replaced by
$W_{\phi}^{(1)}=[w_1+w_2,w_3,w_4,w_5]$
without changing the bias or network output.

The projections
$W_d:\mathbb{R}^{768}\rightarrow\mathbb{R}^{192}$ and
$W_o:\mathbb{R}^{192}\rightarrow\mathbb{R}^{768}$
map the anchor-to-region difference into and out of the hidden space.
The combined hidden representation is normalized by LayerNorm and
followed by SiLU.
The scalar residual gate $g$ is initialized to $-2.0$, and the weight
and bias of $W_o$ are zero-initialized.
Carrier parameters are shared across source views, regions, and
hierarchy levels within each backbone; different backbones use
independently trained Carrier checkpoints.
The spatial token grids are $37\times37$ for LagerNVS,
$32\times32$ for LVSM Decoder-Only, and $32\times56$ for CLiFT.

\paragraph{Training objectives.}
We expand the three training terms in
Eq.~(\ref{eq:carrier_training}).
The rendering objective combines RGB reconstruction and first-order
image-gradient consistency:
\begin{equation}
\begin{aligned}
\mathcal{L}_{\mathrm{render}}
={}&
\lambda_1\,
\operatorname{mean}\!\left|I^S-I^{GT}\right|
+
\lambda_{\mathrm{mse}}\,
\operatorname{mean}\!\left[(I^S-I^{GT})^2\right]
\\
&+
\lambda_{\mathrm{grad}}
\Big(
\operatorname{mean}\!\left|
\nabla_x I^S-\nabla_x I^{GT}
\right|
+
\operatorname{mean}\!\left|
\nabla_y I^S-\nabla_y I^{GT}
\right|
\Big).
\end{aligned}
\label{eq:app_render}
\end{equation}
Here, $\nabla_x$ and $\nabla_y$ denote first-order differences between
neighboring pixels. All reductions average over the corresponding
batch, target, channel, and spatial dimensions.

RGB distillation is applied independently to each scene--target pair
when the teacher has lower reconstruction error than the student.
Let $\mathcal{B}$ and $\mathcal{T}$ denote the batch and target index sets,
and let $N_{\mathrm{rgb}}$ be the number of scalar values in one RGB image.
We define
$e^S_{b,t}=\|I^S_{b,t}-I^{GT}_{b,t}\|_1/N_{\mathrm{rgb}}$
and
$e^T_{b,t}=\|I^T_{b,t}-I^{GT}_{b,t}\|_1/N_{\mathrm{rgb}}$.
The quality-gated distillation loss is
\begin{equation}
g_{b,t}
=
\mathbf{1}\!\left[e^T_{b,t}<e^S_{b,t}\right],
\qquad
\mathcal{L}_{\mathrm{distill}}
=
\frac{1}{|\mathcal{B}||\mathcal{T}|}
\sum_{b\in\mathcal{B}}
\sum_{t\in\mathcal{T}}
g_{b,t}
\frac{
\|I^S_{b,t}-I^T_{b,t}\|_1
}{
N_{\mathrm{rgb}}
}.
\label{eq:app_distill}
\end{equation}
The student error used to compute $g_{b,t}$ is detached.
Inactive terms remain in the full batch--target normalization.

Carrier regularization constrains active non-leaf Carriers toward
their corresponding regional feature means.
Let $\mathcal{A}_{t,i}$ denote the active non-leaf regions of source
view $i$ under the quota for target $t$, and let
$\mathcal{T}_{\mathrm{reg}}$ denote the training targets used for
regularization.
The valid target--view pairs are
\begin{equation}
\mathcal{J}
=
\left\{
(t,i)
\;\middle|\;
t\in\mathcal{T}_{\mathrm{reg}},
\;
\mathcal{A}_{t,i}\neq\varnothing
\right\}.
\label{eq:app_valid_pairs}
\end{equation}
The regularization loss is
\begin{equation}
\mathcal{L}_{\mathrm{carrier}}
=
\frac{1}{|\mathcal{J}|}
\sum_{(t,i)\in\mathcal{J}}
\frac{1}{|\mathcal{A}_{t,i}|C}
\sum_{R\in\mathcal{A}_{t,i}}
\sum_{c=1}^{C}
\operatorname{SmoothL1}
\left(c_{R,c},\mu_{R,c}\right).
\label{eq:app_carrier}
\end{equation}
SmoothL1 uses $\beta=1$.
Regions are averaged uniformly within each valid target--view pair,
without area-based weighting.
For LagerNVS and LVSM Decoder-Only,
$\mathcal{T}_{\mathrm{reg}}$ contains the first training target;
for CLiFT, it contains both training targets.

\paragraph{Teacher context.}
For all three backbones, the teacher representation
$X^T\subseteq X$ uses a fixed context $\mathcal{C}_{24}$ of
24 pose-covered source views selected by the procedure in
Appendix~\ref{app:view_selection}.
Each selected teacher view retains its complete token grid.

\paragraph{Backbone-specific training and checkpoint selection.}
Only the Carrier module is trained; all pretrained backbone parameters
remain frozen.
Source and target views are disjoint, and each training sample uses
two distinct target views.
The deterministic view-construction procedure and the fixed
$\mathcal{T}_{20}$ target pool are defined in
Appendix~\ref{app:view_selection}.
Each backbone uses its corresponding source-view regime and
decoder-visible budget.

\textbf{LagerNVS.}
Carrier training uses 10,139 DL3DV scenes, with source-view counts
$V\in\{60,80,100,120\}$ and a fixed budget of $A=10$.
Each of the 12 epochs traverses all training scenes using the
nested pose-coverage source contexts.
For each scene, a deterministic permutation of the fixed
$\mathcal{T}_{20}$ pool selects two distinct targets per epoch.
The first ten epochs cover all 20 targets once; subsequent epochs
deterministically revisit a subset.

Validation uses 200 held-out DL3DV scenes and four fixed targets
per source-view setting.
We select the Carrier checkpoint by minimizing the weighted
ground-truth validation reconstruction loss over the settings with
80, 100, and 120 source views, using weights $0.2$, $0.3$, and
$0.5$, respectively.
The selected checkpoint is from epoch 9.

\textbf{LVSM Decoder-Only.}
We use 2,048 RE10K scenes for Carrier training and 128 for validation.
These partitions and the 1,633-scene large-view evaluation set are
mutually disjoint.
The training scenes are assigned to three fixed groups by a
deterministic hash of their scene identifiers, containing
661, 700, and 687 scenes, respectively.
The source-view count assigned to each group is rotated among
$V\in\{80,100,120\}$ across epochs.
Let $\operatorname{HashBucket}(s)\in\{0,1,2\}$ denote the group
of scene $s$, and let $(v_0,v_1,v_2)=(80,100,120)$.
At epoch $e$, the source-view count is
\begin{equation}
V(s,e)
=
v_{(e-1+\operatorname{HashBucket}(s))\bmod 3}.
\label{eq:app_lvsm_view_schedule}
\end{equation}
Over 12 epochs, each scene is therefore used four times at each
source-view count.
Training uses the same deterministic pose-coverage and target-sampling
procedure as LagerNVS, under a fixed $A=10$ budget.

For checkpoint selection, each validation scene uses four fixed
target indices $\{2,7,12,17\}$ from its 20-view target pool.
Evaluation across the three source-view settings yields 1,536
rendered targets per validation round.
We select the Carrier checkpoint by validation PSNR;
the selected checkpoint is from epoch 12.

\textbf{CLiFT.}
Carrier training uses 2,048 DL3DV scenes with source-view counts
$V\in\{24,36,48\}$ and a fixed budget of $A=6$.
Training runs for 24 epochs, totaling 3,072 optimizer steps.
Nested pose-coverage source contexts are constructed within the
first 100 available views after reserving the target pool.
A deterministic scene-hash and epoch-dependent cycle assign the
source-view setting, such that each scene is trained eight times
at each source-view count.
Two distinct targets are selected from opposite halves of the
fixed $\mathcal{T}_{20}$ pool.

Validation uses 60 held-out scenes and all 20 target views
for each source-view setting.
We select the Carrier checkpoint using equal-weight macro PSNR
across the three settings.
The selected checkpoint is from epoch 23.

\paragraph{Optimization.}
We share the loss weights and optimizer hyperparameters across all
three backbones:
$\lambda_1=1$,
$\lambda_{\mathrm{mse}}=0.25$,
$\lambda_{\mathrm{grad}}=0.10$,
$\lambda_{\mathrm{dist}}=0.25$, and
$\lambda_{\mathrm{carrier}}=0.05$.
We optimize the Carrier with AdamW using an initial learning rate of
$1\times10^{-5}$, weight decay of $1\times10^{-4}$, and gradient
clipping at $1.0$.
A 20-step warmup is followed by cosine decay to
$1\times10^{-6}$.

LagerNVS, LVSM Decoder-Only, and CLiFT use a per-GPU batch size of 4
on four GPUs, giving a global batch size of 16.
Gradient accumulation is not used.

\subsection{TACC Inference Procedure}
\label{app:tacc_inference}

\setcounter{algorithm}{0}
\renewcommand{\thealgorithm}{\arabic{algorithm}}

Algorithm~\ref{alg:tacc_inference} specifies the target-conditioned
allocation and frontier readout described in
Sec.~\ref{sec:evidence_allocation}.
Each backbone provides the center-pixel ray directions and geometric
descriptors required for allocation, as detailed in
Appendix~\ref{app:backbone_integration}.
All evaluated settings satisfy $V\geq2$ and $2P\leq B\leq VP$.

The raw geometric distance is
$\tilde{\delta}_i^t=\|\xi_i-\xi_t\|_2$,
which is normalized over the source views as
\begin{equation}
\delta_i^t
=
\frac{
\tilde{\delta}_i^t-\min_j\tilde{\delta}_j^t
}{
\max_j\tilde{\delta}_j^t-\min_j\tilde{\delta}_j^t+\epsilon_\delta
},
\qquad
\delta_i^t\equiv\delta(\xi_i,\xi_t).
\label{eq:app_distance_normalization}
\end{equation}

For the standardization in
Eq.~(\ref{eq:relevance_standardization}), the population standard deviation
is computed as
$\sigma_{\mathcal{S}_t}
=
\sqrt{
|\mathcal{S}_t|^{-1}
\sum_{i\in\mathcal{S}_t}
(r_i^t-\mu_{\mathcal{S}_t})^2
}$.
We use the same $\lambda_d$ and $\tau$ as in
Sec.~\ref{sec:evidence_allocation}, and set
$\epsilon_\delta=10^{-6}$ and $\epsilon_s=10^{-8}$.
The normalized entropy is clipped to $[0,1]$ as a numerical safeguard.

\begin{algorithm}[t]
\caption{Target-Conditioned Allocation and Frontier Readout in TACC.}
\label{alg:tacc_inference}
\begin{algorithmic}

\Statex \textbf{Input:}
Carrier hierarchies $\{H_i\}_{i=1}^{V}$,
source center-pixel ray directions $\{d_i\}_{i=1}^{V}$,
source geometric descriptors $\{\xi_i\}_{i=1}^{V}$,
target center-pixel ray direction $d_t$,
target geometric descriptor $\xi_t$,
budget $B$, and tokens per view $P$.

\Statex \textbf{Output:}
Target-conditioned representation $Z_t$ with $|Z_t|=B$.

\Statex
\Statex \textit{Target relevance and full-view preservation:}

\For{$i=1$ to $V$}
    \State $\tilde{\delta}_i^t \gets \|\xi_i-\xi_t\|_2$
\EndFor

\State Normalize $\{\tilde{\delta}_i^t\}_{i=1}^{V}$ using
Eq.~(\ref{eq:app_distance_normalization})

\State Compute $\{r_i^t\}_{i=1}^{V}$ using
Eq.~(\ref{eq:target_relevance})

\State $q^t \gets \mathbf{0}$

\State $i_t^\star \gets \arg\max_i r_i^t$

\State $q_{i_t^\star}^t \gets P$

\State $\mathcal{S}_t
\gets
\{1,\ldots,V\}\setminus\{i_t^\star\}$

\State $B_{\mathrm{rem}} \gets B-P$

\Statex
\Statex \textit{Entropy-adaptive evidence breadth:}

\State Compute $\{z_i^t\}_{i\in\mathcal{S}_t}$ using
Eq.~(\ref{eq:relevance_standardization})

\State Compute $H^t$ and $q_{\mathrm{floor}}^t$ using
Eq.~(\ref{eq:entropy_floor})

\State $B_{\mathrm{res}}^t
\gets
B_{\mathrm{rem}}
-
|\mathcal{S}_t|q_{\mathrm{floor}}^t$

\Statex
\Statex \textit{Relevance-weighted residual allocation:}

\State $\hat p^t \gets \operatorname{Softmax}(z^t/\tau)$

\For{$i\in\mathcal{S}_t$}
    \State $a_i^t
    \gets
    B_{\mathrm{res}}^t\hat p_i^t$
\EndFor

\State $\{b_i^t\}_{i\in\mathcal{S}_t}
\gets
\operatorname{CALR}
\left(
\{a_i^t\},
\{P-q_{\mathrm{floor}}^t\},
B_{\mathrm{res}}^t,
\{r_i^t\}
\right)$

\For{$i\in\mathcal{S}_t$}
    \State $q_i^t
    \gets
    q_{\mathrm{floor}}^t+b_i^t$
\EndFor

\Statex
\Statex \textit{Frontier readout:}

\State $Z_t
\gets
\operatorname*{Concat}_{i=1}^{V}
\operatorname{Read}(H_i,q_i^t)$

\State \Return $Z_t$

\end{algorithmic}
\end{algorithm}

Algorithm~\ref{alg:calr} details the capacity-aware largest-remainder
allocation used in Algorithm~\ref{alg:tacc_inference}.
It converts soft residual quotas into integer allocations while respecting
per-view capacities and the exact residual budget.
Relevance serves only as a small secondary ranking term.

\begin{algorithm}[t]
\caption{Capacity-Aware Largest-Remainder Allocation.}
\label{alg:calr}
\begin{algorithmic}

\Statex \textbf{Input:}
Soft quotas $\{a_i\}_{i\in\mathcal{S}}$,
capacities $\{c_i\}_{i\in\mathcal{S}}$,
target total $B_{\mathrm{res}}$,
and relevance scores $\{r_i\}_{i\in\mathcal{S}}$.

\Statex \textbf{Output:}
Integer allocations $\{b_i\}_{i\in\mathcal{S}}$
satisfying $\sum_i b_i=B_{\mathrm{res}}$
and $0\leq b_i\leq c_i$.

\For{$i\in\mathcal{S}$}
    \State $b_i
    \gets
    \min(\lfloor a_i\rfloor,c_i)$
\EndFor

\State $R
\gets
B_{\mathrm{res}}
-
\sum_{i\in\mathcal{S}}b_i$

\State $\bar r
\gets
\operatorname{MinMax}
\left(
\{r_i\}_{i\in\mathcal{S}}
\right)$

\For{$i\in\mathcal{S}$}
    \State $s_i
    \gets
    \operatorname{frac}(a_i)
    +
    10^{-6}\bar r_i$
\EndFor

\State $\pi
\gets
\operatorname{ArgsortDescending}
\left(
\{s_i\}_{i\in\mathcal{S}}
\right)$

\While{$R>0$}
    \For{$i\in\pi$}
        \If{$R>0$ and $b_i<c_i$}
            \State $b_i \gets b_i+1$
            \State $R \gets R-1$
        \EndIf
    \EndFor
\EndWhile

\State \Return $\{b_i\}_{i\in\mathcal{S}}$

\end{algorithmic}
\end{algorithm}

\subsection{Backbone Integration Details}
\label{app:backbone_integration}

\paragraph{Camera geometry for allocation.}
All three backbone adapters use the center-pixel ray direction for the
directional term in Eq.~(\ref{eq:target_relevance}).
The geometric descriptor $\xi$ used for the distance term is the
center-ray Pl{\"u}cker moment for LagerNVS, the center-ray origin for
LVSM Decoder-Only, and the camera center from the co-normalized c2w
pose for CLiFT.
For CLiFT, camera normalization is computed separately for each
source-view setting using its source set.
The resulting coordinate transform and scale are applied jointly to
the source and target poses before extracting their camera centers.

\paragraph{LagerNVS.}
For LagerNVS, $X_i$ is the final per-view scene representation produced
by the Reconstructor after camera-conditioned VGGT processing.
Unlike initial image patch embeddings, these tokens already incorporate
cross-view interactions.
TACC is inserted between the Reconstructor and Renderer.
The cached Carrier hierarchies provide $Z_t$ as scene memory for the
unchanged Renderer, while target rays follow the original
target-embedding pathway.

\paragraph{LVSM Decoder-Only.}
For LVSM Decoder-Only, TACC is inserted after source-image tokenization
and before source and target tokens are concatenated for joint
Transformer processing.
Each $X_i$ consists of posed-image patch embeddings derived from the
source RGB image and its Pl{\"u}cker rays, before joint source--target
processing.
TACC operates only on the source-token stream; target Pl{\"u}cker rays
are independently converted into target tokens.
The resulting $Z_t$ is concatenated with the target tokens and passed
through the unchanged joint Transformer, after which only the target-token
outputs are decoded into the rendered image.

\paragraph{CLiFT.}
For CLiFT, $X_i$ is the dense per-view source representation produced
by the source encoder.
Progressive Carrier encoding and target-conditioned frontier readout
replace CLiFT's native clustering-and-NeuralCondenser compression pathway,
rather than introducing an additional compression stage after its
compact representation.
For each target, $Z_t$ serves directly as scene memory for the original
Transformer decoder, while target Pl{\"u}cker rays follow the native
target-embedding pathway.
The subsequent normalization and image reconstruction modules remain
unchanged.
The $A=6$ budget constrains the source scene-memory tokens visible
to each target.

\subsection{View Selection Protocol}
\label{app:view_selection}

We define the target-view sets and nested source-view contexts used in
training and evaluation. The candidate range is the first 180 valid views
for LagerNVS on DL3DV, the first 200 valid views for LagerNVS and LVSM
Decoder-Only on RE10K, and the first 100 valid views for CLiFT on DL3DV.

Let $\mathcal{F}=\{j_0,\ldots,j_{M-1}\}$ denote the ordered valid frames
of a scene. Each stored pose $T_j=T_j^{\mathrm{c2w}}$ is used in
camera-to-world form. The camera center and world-space viewing direction
are extracted from its translation column and negative local $z$-axis:
\begin{equation}
\mathbf{c}_j=T_j[:3,3],
\qquad
\mathbf{f}_j=
\frac{-T_j[:3,2]}
{\|T_j[:3,2]\|_2}.
\label{eq:app_pose_extraction}
\end{equation}
These quantities determine view indices only. Camera normalization and
ray construction are performed after the source and target indices
are fixed.

We assign each frame a one-dimensional cumulative pose coordinate.
Let $\alpha$ be the median non-zero translation distance between
adjacent valid frames. For adjacent frames $j_{k-1}$ and $j_k$, define
\begin{equation}
\delta_k
=
\|\mathbf{c}_{j_k}-\mathbf{c}_{j_{k-1}}\|_2
+
0.1\,\alpha\,
\frac{
\arccos\!\left(
\operatorname{clip}\!\left(
\mathbf{f}_{j_k}^{\mathsf T}\mathbf{f}_{j_{k-1}},
-1,1
\right)
\right)
}{\pi}.
\label{eq:app_pose_increment}
\end{equation}
The normalized cumulative coordinate is
\begin{equation}
s_{j_0}=0,
\qquad
s_{j_k}
=
\frac{
\sum_{\ell=1}^{k}\delta_\ell
}{
\sum_{\ell=1}^{M-1}\delta_\ell
},
\qquad
k=1,\ldots,M-1.
\label{eq:app_pose_coordinate}
\end{equation}

For a view set $\mathcal{U}$ with pose coordinates $\{s_j\}$,
$\operatorname{ProgressiveFPS}(\mathcal{U})$ denotes a deterministic
progressive farthest-point ordering. We first select the views with
the minimum and maximum pose coordinates. Each subsequent view is
selected by
\begin{equation}
j^\star
=
\arg\max_{j\in\mathcal{U}\setminus\mathcal{S}}
\min_{k\in\mathcal{S}}
|s_j-s_k|,
\label{eq:app_progressive_fps}
\end{equation}
where $\mathcal{S}$ is the set of previously selected views.
Ties are resolved by the smaller frame index.

\paragraph{Target selection.}
We construct a 40-view target pool using the reference coordinates
\begin{equation}
u_r=\frac{r+0.5}{40},
\qquad
r=0,\ldots,39.
\label{eq:app_target_reference}
\end{equation}
For each $u_r$, we select the remaining frame that minimizes
$|s_j-u_r|$, resolving ties by the smaller frame index.
The selected frames form $\mathcal{T}_{40}$.

For a required target count $N\leq40$, the standard target set is
\begin{equation}
\mathcal{T}_{N}
=
\operatorname{SortIndex}\!\left(
\operatorname{Prefix}_{N}\!\left(
\operatorname{ProgressiveFPS}(\mathcal{T}_{40})
\right)
\right).
\label{eq:app_target_set}
\end{equation}
Thus, $\mathcal{T}_{20}$ contains the first 20 views in the
progressive farthest-point ordering of $\mathcal{T}_{40}$, sorted by
frame index.

LagerNVS uses $\mathcal{T}_{20}$ for DL3DV evaluation.
The RE10K evaluations of LagerNVS and LVSM Decoder-Only use
$\mathcal{T}_{12}$, while CLiFT uses $\mathcal{T}_{20}$.
Within each scene and evaluation setting, the target set is fixed
across source-view counts.

\paragraph{Source-view selection.}
We exclude the entire target pool $\mathcal{T}_{40}$ from the source
candidates and construct a single progressive ordering:
\begin{equation}
\mathcal{O}_{s}
=
\operatorname{ProgressiveFPS}
\left(
\mathcal{F}\setminus\mathcal{T}_{40}
\right).
\label{eq:app_source_order}
\end{equation}
For a required source-view count $K$, the source context is
\begin{equation}
\mathcal{C}_{K}
=
\operatorname{SortIndex}\!\left(
\operatorname{Prefix}_{K}(\mathcal{O}_{s})
\right).
\label{eq:app_source_context}
\end{equation}
All source contexts are prefixes of the same ordering and are
therefore nested:
\begin{equation}
K_1<K_2
\quad\Longrightarrow\quad
\mathcal{C}_{K_1}\subset\mathcal{C}_{K_2}.
\label{eq:app_nested_contexts}
\end{equation}

Since the standard target sets are subsets of $\mathcal{T}_{40}$
and all source views belong to
$\mathcal{F}\setminus\mathcal{T}_{40}$, source and target views are
strictly disjoint. The target-count experiment in
Appendix~\ref{app:runtime} uses a separate target-selection protocol
defined there rather than Eq.~(\ref{eq:app_target_set}).

\begin{table}[H]
    \centering
    \caption{
        \textbf{Baseline implementation and exact-budget matching.}
        Each compressed baseline exposes exactly $B=A\cdot P$
        source-side tokens.
    }
    \label{tab:app_baselines}

    \small
    \setlength{\tabcolsep}{4pt}
    \renewcommand{\arraystretch}{1.10}

    \begin{tabular}{
        @{}
        p{0.16\linewidth}
        p{0.78\linewidth}
        @{}
    }
        \toprule

        \textbf{Method}
        & \textbf{Exact-budget realization} \\

        \midrule

        ToMe
        &
        Each source view receives $\lfloor B/V\rfloor$ tokens,
        with one additional token assigned to each of the first
        $B\bmod V$ views. BSM and mass-weighted merging are applied
        within each view until its assigned quota is reached. \\

        DivPrune
        &
        Greedy max--min diversity selection is applied jointly to
        all source tokens. Exactly $B$ tokens are retained and
        restored to their original sequence order before rendering. \\

        ZPressor
        &
        $A=B/P$ anchor views are used, each producing a complete
        $P$-token grid. The resulting representation contains
        exactly $AP=B$ source tokens. \\

        \bottomrule
    \end{tabular}

\end{table}

\subsection{Baseline Implementation}
\label{app:baseline}

For each backbone, TACC and the adapted compression baselines operate
on the same source representation $X$ defined in
Sec.~\ref{sec:overview}.
The decoder-visible budget $B=A\cdot P$ counts source-side tokens only.
Target-query, target-ray, target-pose, and backbone-internal register
tokens remain unchanged and are excluded from budget matching.
ToMe and DivPrune are training-free; the training protocols for
CLiFT Native and ZPressor are specified below. Tab.~\ref{tab:app_baselines} summarizes the exact-budget
implementation of the adapted compression baselines.

\paragraph{CLiFT Native training protocol.}
For the controlled comparison in Tab.~\ref{tab:clift_scaling},
we retrain CLiFT's native compression pathway on the same 2,048
DL3DV scenes as TACC. Both methods use the same source-view schedule
($V\in\{24,36,48\}$), target-sampling protocol, and fixed $A=6$
decoder-visible budget. Training runs for 24 epochs, totaling
3,072 optimizer steps, with a global batch size of 16 across
four GPUs.

Both methods share the same pretrained source encoder and renderer,
which remain frozen throughout training. Only their respective
compression modules are optimized, using method-specific training
objectives and optimization settings. Validation uses the same
60 held-out scenes and all 20 target views per source-view setting;
checkpoint selection uses equal-weight macro PSNR across the three
settings. Unlike CLiFT's official second-stage training, the renderer
is not fine-tuned in this controlled comparison.

\paragraph{ZPressor training protocol.}
For each backbone, ZPressor uses the same training split,
source-view schedule, target-sampling protocol, and validation
split as TACC.
Only the compressor is trained; the pretrained NVS backbone remains
frozen. We retain the ZPressor training objective and optimization
configuration.

\textbf{LagerNVS.}
ZPressor is trained on 10,139 DL3DV scenes with source-view counts
$V\in\{60,80,100,120\}$ and a fixed decoder-visible budget of $A=10$.
Validation uses 200 held-out scenes and four fixed targets per
source-view setting.
Checkpoint selection minimizes the weighted validation reconstruction
objective over the settings with 80, 100, and 120 source views,
using weights $0.2$, $0.3$, and $0.5$, respectively.
The compressor is evaluated every 2,500 optimization steps, and the
checkpoint at step 25,000 is selected.
For the tighter-budget evaluation in
Appendix~\ref{app:tighter_budget}, the compressor trained with an $A=10$
budget is evaluated at $A=6$ without budget-specific retraining.

\textbf{LVSM Decoder-Only.}
ZPressor is trained on 2,048 RE10K scenes with source-view counts
$V\in\{80,100,120\}$ and $A=10$.
Training targets follow the LVSM protocol in
Appendix~\ref{app:carrier_training}.
Validation uses 128 held-out RE10K scenes and the fixed target
indices $\{2,7,12,17\}$ under all three source-view settings;
the checkpoint is selected by validation PSNR.

\textbf{CLiFT.}
ZPressor is trained on the same DL3DV training split as TACC,
with source-view counts $V\in\{24,36,48\}$ and $A=6$.
Training targets follow the CLiFT protocol in
Appendix~\ref{app:carrier_training}.
Validation uses 60 held-out scenes and all 20 target views per
source-view setting; the checkpoint is selected by equal-weight
macro PSNR across the three settings.

All ZPressor compressors are trained for 25,000 optimization steps
using AdamW with an initial learning rate of $1\times10^{-4}$,
$\beta_1=0.9$, $\beta_2=0.95$, and weight decay $0.05$.
A 1,000-step linear warmup is followed by cosine decay to
$1\times10^{-5}$.

LagerNVS and LVSM Decoder-Only use a per-GPU batch size of 4
on four GPUs, giving a global batch size of 16.
CLiFT uses a per-GPU batch size of 1 on four GPUs, giving a
global batch size of 4.

\section{Additional Results}
\label{app:additional_results}

\subsection{Additional Evaluation on LVSM Decoder-Only}
\label{app:lvsm_dense_scaling}

We extend the LVSM Decoder-Only evaluation to 15--120 source views
using the same 1,633-scene RE10K test set as in the main experiments.
Each scene is restricted to its first 200 valid views, with 12 target
views evaluated at $256\times256$ resolution and a target-view batch
size of 4.
Full Context and TACC use identical source and target views, which
are disjoint by construction.
TACC uses a fixed $A=10$ decoder-visible budget throughout this
evaluation and reuses the Carrier checkpoint trained with
$V\in\{80,100,120\}$ without source-count-specific fine-tuning.

\begin{figure}[H]
    \centering
    \includegraphics[width=\linewidth]
    {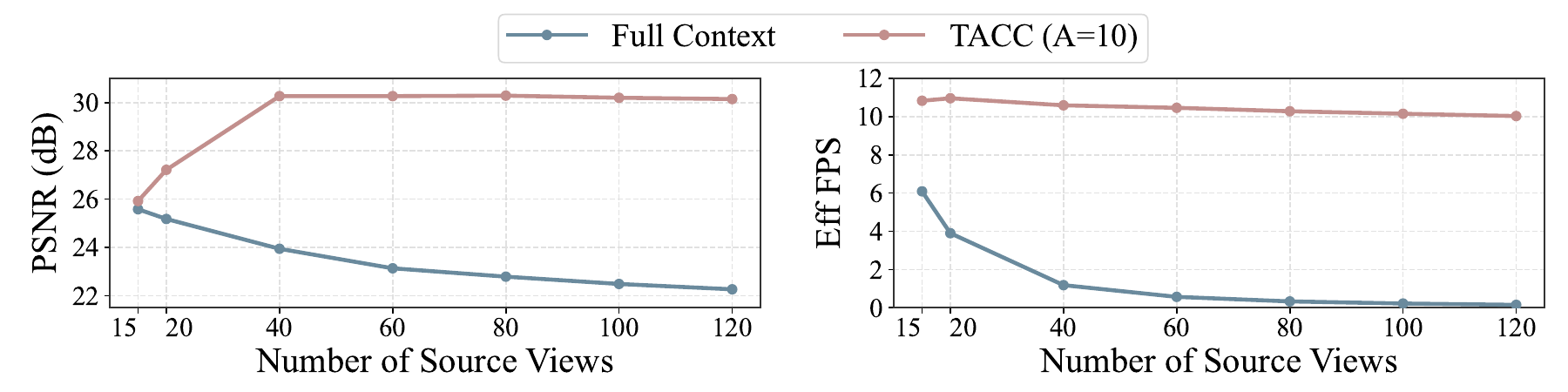}
    \caption{
        \textbf{LVSM Decoder-Only with increasing source views.}
        PSNR (left) and Effective FPS (right) on RE10K across
        15--120 source views.
        TACC uses a fixed $A=10$ decoder-visible budget, whereas
        Full Context processes all selected source tokens.
    }
    \label{fig:lvsm_dense_scaling}
\end{figure}

As shown in Fig.~\ref{fig:lvsm_dense_scaling}, TACC's reconstruction
quality increases up to 40 source views and remains nearly stable
thereafter, with little variation in Effective FPS.
In contrast, the frozen Full Context model shows decreasing
reconstruction quality and target-time throughput as its source
context grows.
At 120 source views, TACC reaches 30.14~dB PSNR, compared with
22.26~dB for Full Context, while maintaining the fixed
decoder-visible budget.

The Full Context results reflect a substantial shift from the
pretrained LVSM setting: the scene-level backbone is trained
with 2 source views~\citep{lvsm}, whereas the 120-view evaluation
uses $60\times$ more source views and exposes 122,880 source patch
tokens at $256\times256$ resolution.
This larger context increases rendering cost and may contribute
to the observed quality degradation.
Related observations in VGGT suggest that additional redundant
views may dilute informative geometric evidence rather than
consistently improve reconstruction quality
~\citep{park2026diversityawareviewpartitioningscalable}.

The Full Context reference in Fig.~\ref{fig:lvsm_dense_scaling}
therefore represents the frozen pretrained backbone; a separate
comparison with dense-context fine-tuning is provided below.

\paragraph{Carrier storage.}
For LVSM Decoder-Only, the cached node-Carriers form a BF16 tensor
of shape $V\times(2P-1)\times C$, where $P=1024$ and $C=768$.
At 120 source views, this tensor occupies 359.82~MiB.
This calculated footprint excludes other retained tensors, hierarchy
metadata, and temporary allocations; it does not represent the total
Carrier cache or incremental peak GPU-memory overhead.
The Carrier hierarchies are constructed once per source set and
reused across target views.

\paragraph{Stage-wise peak GPU memory.}
To further examine the memory scaling in
Fig.~\ref{fig:lvsm_efficiency}(a), we profile peak GPU memory
during source tokenization, Carrier construction, and decoding
under the same LVSM evaluation setting. Stage-wise peaks include
tensors retained from preceding stages, and the overall peak is
the maximum across stages.

\begin{table}[!htbp]
    \centering
    {%
    \caption{
        \textbf{Stage-wise peak GPU memory on RE10K.}
        All values are in GiB.
        Source tokenization is identical for Full Context and TACC.
        Build denotes TACC's one-time Carrier construction.
        Overall peaks include retained tensors from preceding stages.
    }
    \label{tab:app_lvsm_stage_memory}
    \small
    \setlength{\tabcolsep}{3.5pt}
    \renewcommand{\arraystretch}{1.08}
    \begin{tabular*}{\linewidth}{@{}c@{\extracolsep{\fill}}cccccc@{}}
        \toprule
        \textbf{Views}
        & \textbf{Source Token.}
        & \textbf{Full Decode}
        & \textbf{TACC Decode}
        & \textbf{TACC Build}
        & \textbf{Full Peak}
        & \textbf{TACC Peak} \\
        \midrule
        15  & 0.82 & 2.31 & 2.11 & 1.12 & 2.31  & 2.11 \\
        20  & 0.86 & 2.81 & 2.17 & 1.27 & 2.81  & 2.17 \\
        40  & 1.05 & 4.80 & 2.40 & 1.86 & 4.80  & 2.40 \\
        60  & 1.23 & 6.79 & 2.63 & 2.45 & 6.79  & 2.63 \\
        80  & 1.41 & 8.77 & 2.83 & 3.03 & 8.77  & 3.03 \\
        100 & 1.59 & 10.76 & 3.06 & 3.62 & 10.76 & 3.62 \\
        120 & 1.78 & 12.77 & 3.34 & 4.22 & 12.77 & 4.22 \\
        \bottomrule
    \end{tabular*}
    }
\end{table}

As shown in Tab.~\ref{tab:app_lvsm_stage_memory},
source tokenization incurs identical memory costs for both
methods, while the primary difference arises during decoding.
At 120 source views, TACC reduces the decoder-stage peak
from 12.77 to 3.34~GiB by limiting the decoder-visible
source representation to the fixed $A=10$ budget.
Its overall peak increases from 2.11 to 4.22~GiB across
15--120 source views, compared with 2.31 to 12.77~GiB
for Full Context. At 80--120 source views, TACC's overall
peak occurs during the one-time Carrier construction rather
than decoding. This construction cost is distinct from the
cached node-Carrier storage reported above. Thus, the fixed
decoder-visible budget limits rendering-stage memory growth,
while hierarchy construction introduces a separate
source-dependent memory cost.

\paragraph{Target-aware Top-$A$ control.}
To examine whether selecting relevant complete views alone accounts
for TACC's reconstruction quality, we evaluate a Top-$A$ Full Views
control at 120 source views.
For each target, this control retains the ten highest-ranked source
views according to TACC's geometric relevance score, without
Carrier compression or cross-view token allocation.
Both methods expose exactly $A=10$ complete-view equivalents
to the decoder.

As shown in Tab.~\ref{tab:lvsm_topa_control}, TACC substantially improves
reconstruction quality over Top-$A$ Full Views while maintaining comparable
target-time throughput. This result shows that selecting target-relevant
complete views alone is insufficient to match TACC under the same
decoder-visible budget.

\begin{table}[!htbp]
    \centering
    \caption{
        \textbf{Target-aware Top-$A$ control on LVSM Decoder-Only
        at 120 source views.}
        Both methods use the same $A=10$ decoder-visible budget.
    }
    \label{tab:lvsm_topa_control}
    \small
    \setlength{\tabcolsep}{6.0pt}
    \renewcommand{\arraystretch}{1.08}
    \begin{tabular}{@{}lcccc@{}}
        \toprule
        \textbf{Method}
        & \textbf{PSNR} $\uparrow$
        & \textbf{SSIM} $\uparrow$
        & \textbf{LPIPS} $\downarrow$
        & \textbf{Eff. FPS} $\uparrow$ \\
        \midrule
        Top-$A$ Full Views
        & 26.7087 & 0.8517 & 0.1154 & \textbf{11.11} \\
        \textbf{TACC}
        & \textbf{30.1441} & \textbf{0.9243} & \textbf{0.0750} & 10.03 \\
        \bottomrule
    \end{tabular}
\end{table}


\paragraph{{Dense-context adaptation of LVSM Decoder-Only.}}

To examine whether the reconstruction-quality advantage over the
released Full Context checkpoint persists after dense-context
adaptation, we additionally fine-tune the full LVSM Decoder-Only
backbone on RE10K using dense source-view inputs.
The fine-tuned model has approximately 170.8M trainable parameters,
and its training and evaluation scenes are strictly disjoint.
We evaluate it on the same 1,633-scene RE10K benchmark using
identical source and target views, with 80, 100, and 120 source
views.
The fine-tuned LVSM processes the complete source-token context,
whereas TACC retains the original frozen backbone and trains only
302,785 Carrier parameters, with a fixed $A=10$
decoder-visible budget.

\begin{table}[!htbp]
    \centering
    
    \caption{
        \textbf{Comparison with dense-context-fine-tuned
        LVSM Decoder-Only on RE10K.}
        The fine-tuned LVSM processes all source tokens,
        whereas TACC retains the frozen pretrained backbone
        with a fixed $A=10$ decoder-visible budget.
        Effective FPS excludes one-time source encoding and
        Carrier construction.
    }
    \label{tab:lvsm_dense_finetuned}
    \small
    \setlength{\tabcolsep}{3.5pt}
    \renewcommand{\arraystretch}{1.08}
    \begin{tabular*}{\linewidth}{@{}c@{\extracolsep{\fill}}cccc@{}}
        \toprule
        \textbf{Source}
        & \multicolumn{2}{c}{\textbf{Fine-tuned LVSM}}
        & \multicolumn{2}{c}{\textbf{TACC}} \\
        \cmidrule(lr){2-3}
        \cmidrule(l){4-5}
        \textbf{Views}
        & \textbf{PSNR} $\uparrow$
        & \textbf{Eff. FPS} $\uparrow$
        & \textbf{PSNR} $\uparrow$
        & \textbf{Eff. FPS} $\uparrow$ \\
        \midrule
        80
        & 29.5903 & 0.3377 & \textbf{30.2851} & \textbf{10.28} \\
        100
        & 29.4271 & 0.2211 & \textbf{30.1978} & \textbf{10.15} \\
        120
        & 29.2294 & 0.1547 & \textbf{30.1441} & \textbf{10.03} \\
        \bottomrule
    \end{tabular*}
\end{table}

As shown in Tab.~\ref{tab:lvsm_dense_finetuned},
dense-context fine-tuning substantially improves LVSM's
reconstruction quality, narrowing the gap observed with
the released frozen checkpoint.
Nevertheless, TACC maintains slightly higher PSNR across
all three source-view settings while delivering substantially
greater target-time rendering throughput.
These results indicate that the large quality gap against
the original Full Context checkpoint should not be
attributed entirely to the compression mechanism.
Rather, TACC enables the recovery or improvement of
dense-context reconstruction quality at substantially
lower target-time rendering cost, without fine-tuning
the original backbone.

\subsection{Evaluation under a Tighter Decoder-Visible Budget}
\label{app:tighter_budget}

We further evaluate LagerNVS on DL3DV with 60, 80, and 100 source views under
a tighter $A=6$ decoder-visible budget. TACC reuses the Carrier checkpoint
trained with an $A=10$ budget without budget-specific fine-tuning; the ZPressor
checkpoint follows the protocol in Appendix~\ref{app:baseline}.

As shown in Tab.~\ref{tab:app_a6}, TACC achieves the highest
reconstruction quality among compressed methods across all three
source-view settings.
At 60 and 80 source views, it also exceeds the frozen Full Context
checkpoint on all three reconstruction metrics.
At 100 source views, where the source-to-budget ratio reaches
approximately $16.7\times$, Full Context retains a slightly higher
PSNR, while TACC achieves higher SSIM and lower LPIPS than all
compared methods.
TACC maintains Effective FPS near 40 across the three settings.

\begin{table}[!htbp]
    \centering
    \caption{
        \textbf{Comparison on DL3DV with a tighter compression budget.}
        Compressed methods use a fixed $A=6$ decoder-visible budget,
        while \colorbox{gray!18}{Full Context} uses the complete
        LagerNVS representation.
    }
    \label{tab:app_a6}

    \small
    \setlength{\tabcolsep}{3.0pt}
    \renewcommand{\arraystretch}{1.08}

    \resizebox{\linewidth}{!}{
    \begin{tabular}{@{}lcccccccccccc@{}}

        \toprule

        &
        \multicolumn{4}{c}{\textbf{60 Source Views}}
        &
        \multicolumn{4}{c}{\textbf{80 Source Views}}
        &
        \multicolumn{4}{c}{\textbf{100 Source Views}}
        \\

        \cmidrule(lr){2-5}
        \cmidrule(lr){6-9}
        \cmidrule(lr){10-13}

        \textbf{Method}
        & PSNR $\uparrow$
        & SSIM $\uparrow$
        & LPIPS $\downarrow$
        & Eff. FPS $\uparrow$

        & PSNR $\uparrow$
        & SSIM $\uparrow$
        & LPIPS $\downarrow$
        & Eff. FPS $\uparrow$

        & PSNR $\uparrow$
        & SSIM $\uparrow$
        & LPIPS $\downarrow$
        & Eff. FPS $\uparrow$
        \\

        \midrule
        \multicolumn{13}{c}{\textbf{LagerNVS-based}} \\
        \midrule

        \rowcolor{gray!18}
        Full Context
        & 26.2985
        & 0.8247
        & 0.0892
        & 5.87

        & 25.6766
        & 0.8064
        & 0.0963
        & 4.40

        & 25.2932
        & 0.7953
        & 0.1012
        & 3.57
        \\
        \hdashline

        ToMe
        & 19.5020
        & 0.5217
        & 0.3859
        & 47.93

        & 18.8116
        & 0.4944
        & 0.4265
        & 47.23

        & 18.6572
        & 0.4877
        & 0.4408
        & 47.61
        \\

        DivPrune
        & 21.1870
        & 0.6289
        & 0.2364
        & 47.97

        & 21.0042
        & 0.6201
        & 0.2429
        & 47.20

        & 20.8862
        & 0.6148
        & 0.2472
        & 47.93
        \\

        ZPressor
        & 23.7222
        & 0.7653
        & 0.1638
        & 47.61

        & 23.5503
        & 0.7589
        & 0.1664
        & 47.93

        & 23.4514
        & 0.7551
        & 0.1671
        & 47.58
        \\

        \textbf{Ours}
        & 27.1015
        & 0.8623
        & 0.0819
        & 40.40

        & 26.6201
        & 0.8504
        & 0.0827
        & 39.52

        & 25.0192
        & 0.8040
        & 0.0954
        & 40.01
        \\

        \bottomrule

    \end{tabular}
    }
\end{table}

\paragraph{Source-view scaling across budgets.}We further evaluate TACC on the official 140-scene
DL3DV benchmark with 20 target views per scene, varying the number of
source views from 20 to 100 under fixed decoder-visible budgets of
$A=6$ and $A=10$. Both settings use the same trained Carrier checkpoint,
without budget-specific retraining.

\begin{table}[!htbp]
    \centering
    {%
    \caption{
        \textbf{Source-view scaling across decoder-visible budgets
        on DL3DV.}
        TACC uses the same Carrier checkpoint for $A=6$
        and $A=10$, without budget-specific retraining.
        Effective FPS excludes one-time source encoding
        and Carrier construction.
    }
    \label{tab:app_budget_scaling}

    \small
    \setlength{\tabcolsep}{2.5pt}
    \renewcommand{\arraystretch}{1.08}

    \begin{tabular*}{\linewidth}{
        @{}c@{\extracolsep{\fill}}cccccccc@{}
    }
        \toprule

        &
        \multicolumn{4}{c}{\textbf{$A=6$}}
        &
        \multicolumn{4}{c}{\textbf{$A=10$}}
        \\

        \cmidrule(lr){2-5}
        \cmidrule(lr){6-9}

        \textbf{Source Views}
        & PSNR $\uparrow$
        & SSIM $\uparrow$
        & LPIPS $\downarrow$
        & Eff. FPS $\uparrow$

        & PSNR $\uparrow$
        & SSIM $\uparrow$
        & LPIPS $\downarrow$
        & Eff. FPS $\uparrow$
        \\

        \midrule

        20
        & 26.6980
        & 0.8513
        & 0.0958
        & 42.37

        & 27.2189
        & 0.8617
        & 0.0893
        & 29.15
        \\

        40
        & 27.0155
        & 0.8595
        & 0.0871
        & 40.41

        & 27.4253
        & 0.8682
        & 0.0834
        & 28.08
        \\

        60
        & \textbf{27.1015}
        & 0.8623
        & 0.0819
        & 40.40

        & \textbf{27.5815}
        & 0.8691
        & 0.0792
        & 28.02
        \\

        80
        & 26.6201
        & 0.8504
        & 0.0827
        & 39.52

        & 27.3415
        & 0.8649
        & 0.0791
        & 27.95
        \\

        100
        & 25.0192
        & 0.8040
        & 0.0954
        & 40.01

        & 26.8787
        & 0.8523
        & 0.0833
        & 28.12
        \\

        \bottomrule
    \end{tabular*}
    }
\end{table}

As shown in Tab.~\ref{tab:app_budget_scaling},
both budgets achieve their highest PSNR at 60 source views, followed
by a decline as the source context grows. From 60 to 100 source views,
PSNR decreases by 2.08 dB at $A=6$, compared with 0.70 dB at $A=10$.
The larger budget also consistently improves SSIM and LPIPS, albeit at
lower Effective FPS. These results suggest that increasing the
decoder-visible budget can mitigate reconstruction degradation under
denser source observations, while introducing a quality--efficiency
trade-off.

\paragraph{Quality--efficiency across budgets.}We compare
TACC, ToMe, DivPrune, and ZPressor under six matched decoder-visible
budgets, $A\in\{2,4,6,8,10,13\}$, using LagerNVS on the official
140-scene DL3DV benchmark with 80 source views and 20 target views
per scene. All methods use identical source and target views and a
target-view batch size of 4. TACC and ZPressor each reuse a single
trained checkpoint across all six budgets, without budget-specific
retraining; ToMe and DivPrune are training-free. The resulting
quality--efficiency curves are shown in
Fig.~\ref{fig:tacc_quality_efficiency}(a), with complete quantitative
results reported in Tab.~\ref{tab:app_budget_tradeoff}.

\begin{table}[!htbp]
    \centering
    {%
    \caption{
        \textbf{Quality--efficiency comparison across budgets on DL3DV.}
        LagerNVS is evaluated with 80 source views and
        matched decoder-visible budgets.
        Each entry reports PSNR (dB) / Effective FPS.
        Trained compressors reuse their respective
        checkpoints across all budgets.
    }
    \label{tab:app_budget_tradeoff}

    \small
    \setlength{\tabcolsep}{3.5pt}
    \renewcommand{\arraystretch}{1.10}

    \begin{tabular*}{\linewidth}{
        @{}c@{\extracolsep{\fill}}cccc@{}
    }
        \toprule

        \textbf{Budget}
        & \textbf{TACC}
        & \textbf{ToMe}
        & \textbf{DivPrune}
        & \textbf{ZPressor}
        \\

        \midrule

        $A=2$
        & \textbf{21.1599} / 71.86
        & 17.0749 / 99.00
        & 19.0492 / 99.16
        & 18.6672 / 101.81
        \\

        $A=4$
        & \textbf{24.3268} / 52.41
        & 18.2611 / 65.08
        & 20.3359 / 64.45
        & 21.4607 / 65.66
        \\

        $A=6$
        & \textbf{26.6201} / 39.52
        & 18.8116 / 47.23
        & 21.0042 / 47.20
        & 23.5503 / 47.93
        \\

        $A=8$
        & \textbf{27.1159} / 32.88
        & 19.6274 / 38.09
        & 21.4596 / 37.29
        & 23.5808 / 38.15
        \\

        $A=10$
        & \textbf{27.3415} / 27.95
        & 20.0344 / 31.43
        & 21.7624 / 31.54
        & 24.1568 / 31.26
        \\

        $A=13$
        & \textbf{27.4844} / 22.47
        & 20.5562 / 24.94
        & 22.1158 / 24.93
        & 24.6789 / 25.09
        \\

        \bottomrule
    \end{tabular*}
    }
\end{table}

As shown in Tab.~\ref{tab:app_budget_tradeoff},
TACC achieves the highest PSNR at every matched budget, exceeding
the strongest competing method by 2.11--3.54 dB. At $A=6$, TACC
obtains 26.62 dB and 39.52 Effective FPS, outperforming ZPressor
at $A=13$ (24.68 dB and 25.09 Effective FPS) in both metrics.
These results illustrate TACC's favorable quality--efficiency
trade-off across compression budgets.

\begin{figure}[H]

    \centering
    \includegraphics[width=\linewidth]{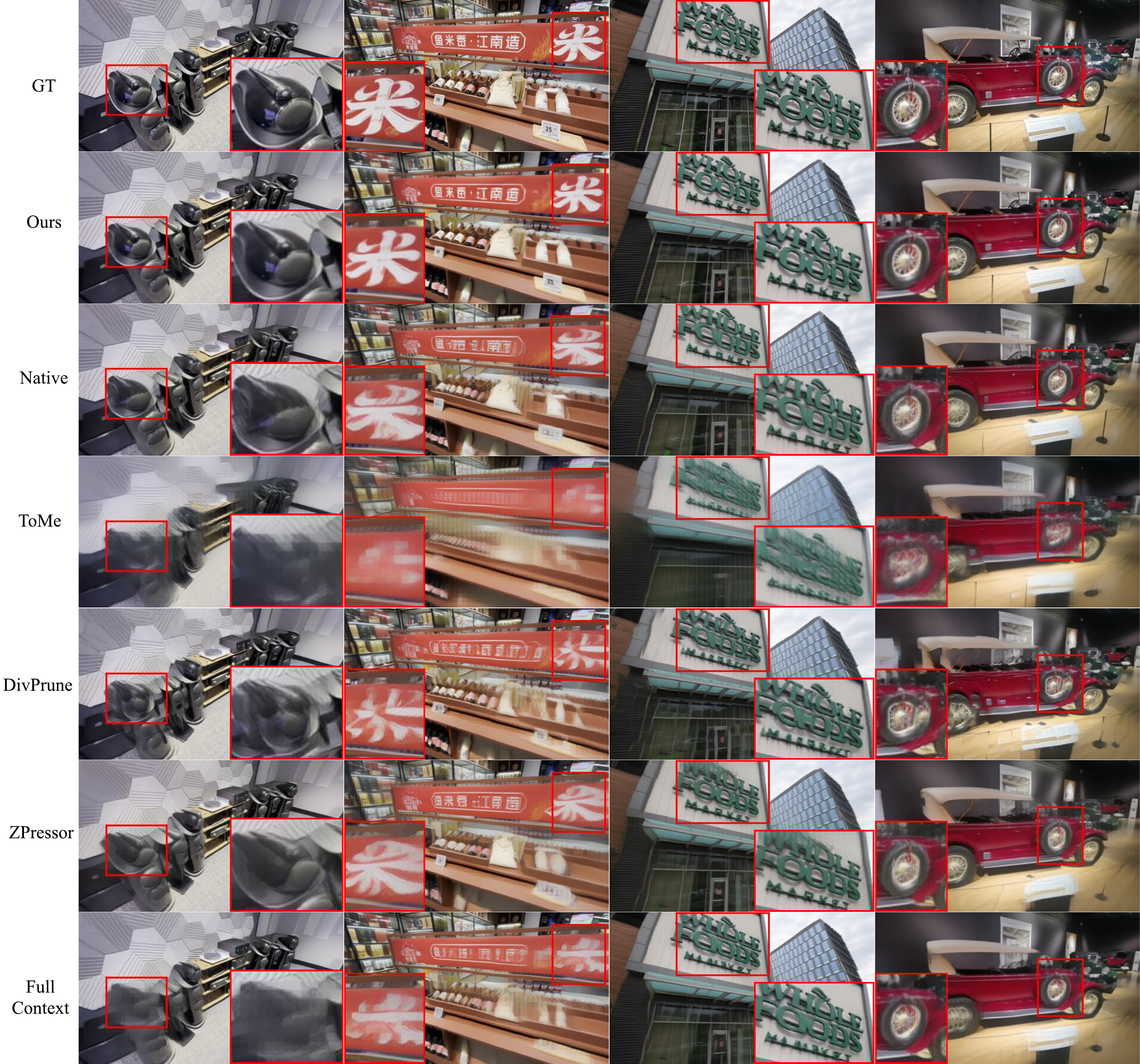}
    \vspace{-5pt}
    \caption{
        \textbf{Additional qualitative results on CLiFT with DL3DV.}
        Native uses the checkpoint retrained under the matched setting
        described in Appendix~\ref{app:baseline}.
        Compressed methods use a fixed $A=6$ decoder-visible budget,
        while Full Context uses all source tokens.
        TACC better preserves local structures and fine-grained details.
    }
    \label{fig:app_clift_qualitative}
\end{figure}

\subsection{Additional Results on CLiFT}
\label{app:clift_qualitative}
\label{app:clift_official}

We additionally evaluate CLiFT's official pretrained Native checkpoint
under the same DL3DV evaluation protocol as
Tab.~\ref{tab:clift_scaling}.
Tab.~\ref{tab:clift_official} reports its reconstruction quality and
target-time throughput across 24, 36, and 48 source views.
These results provide a reference for the released CLiFT model.
Unlike the Native baseline in Tab.~\ref{tab:clift_scaling}, the
official checkpoint is not retrained under our matched
compression-module training protocol.

\begin{table}[H]
    \centering
    \caption{
        \textbf{Evaluation of the official CLiFT Native checkpoint on DL3DV.}
        Results use the same source and target views as
        Tab.~\ref{tab:clift_scaling}, with a fixed $A=6$
        decoder-visible budget.
    }
    \label{tab:clift_official}

    \small
    \setlength{\tabcolsep}{6pt}
    \renewcommand{\arraystretch}{1.10}

    \begin{tabular}{@{}ccccc@{}}
        \toprule
        \textbf{Source Views}
        & \textbf{PSNR $\uparrow$}
        & \textbf{SSIM $\uparrow$}
        & \textbf{LPIPS $\downarrow$}
        & \textbf{Eff. FPS $\uparrow$} \\
        \midrule
        24 & 19.9137 & 0.5764 & 0.3070 & 47.33 \\
        36 & 19.8713 & 0.5719 & 0.3137 & 47.46 \\
        48 & 19.8664 & 0.5718 & 0.3142 & 47.08 \\
        \bottomrule
    \end{tabular}
\end{table}

Fig.~\ref{fig:app_clift_qualitative} provides complementary qualitative
results for the CLiFT comparison.

\subsection{Additional Ablations and Controls}
\label{app:allocation_ablations}

\paragraph{Carrier and allocation controls.}
We conduct additional experiments using LagerNVS on DL3DV with 80 source
views and a fixed $A=10$ decoder-visible budget.
In Tab.~\ref{tab:component_ablation}, \emph{w/o Carrier} retains
TACC's per-view quotas but replaces Carrier readout with approximately
uniform spatial subsampling of the original tokens.
\emph{Uniform Alloc.} instead uses approximately uniform, exact-budget
quotas during both training and evaluation; its Carrier is retrained
following the protocol in Appendix~\ref{app:carrier_training}.

We further evaluate \emph{Residual PMA + Alloc.}, which replaces the
learned regional Carriers with residual Pooling by Multihead Attention
(PMA) aggregation~\citep{settransformer} while retaining TACC's
target-conditioned per-view allocation.
The PMA module is independently trained under the same training and
evaluation protocol as TACC, using identical rendering and quality-gated
RGB distillation losses and weights, but without the Carrier-specific
reconstruction regularizer.

\paragraph{Relevance-score components.}
We isolate the directional and distance terms in
Eq.~(\ref{eq:target_relevance}) on the official 140-scene DL3DV
benchmark, using LagerNVS with 80 source views, 20 target views,
and $A=10$.
\emph{Direction Only} uses
$r_i^t=\langle\hat d_i,\hat d_t\rangle$,
whereas \emph{Moment Only} uses
$r_i^t=-\delta(m_i,m_t)$.
All other inference components remain unchanged.

\begingroup

\paragraph{Allocation safeguards.}
We examine the contributions of Top-1 full-view preservation
and the entropy-based coverage floor using LagerNVS on the
official 140-scene DL3DV benchmark, with 80 source views,
20 target views per scene, and a fixed $A=10$ budget.
Both ablations use the same trained Carrier checkpoint
as full TACC, modifying only the corresponding
inference-time allocation mechanism.

\par
\endgroup

\begin{table}[t]
    \centering
    {%
    \caption{
        \textbf{Additional component ablations on DL3DV.}
        All variants use LagerNVS with 80 source views
        and a fixed $A=10$ decoder-visible budget.
        Residual PMA is independently trained, whereas
        relevance-score and allocation-safeguard ablations
        use the fixed TACC Carrier checkpoint.
    }

    \label{tab:controlled_ablations}
    \label{tab:allocation_ablations}
    \label{tab:app_allocation_safeguards}

    \small
    \setlength{\tabcolsep}{6.5pt}
    \renewcommand{\arraystretch}{1.08}

    \begin{tabular}{@{}lccc@{}}
        \toprule

        \textbf{Variant}
        & \textbf{PSNR} $\uparrow$
        & \textbf{SSIM} $\uparrow$
        & \textbf{LPIPS} $\downarrow$
        \\

        \midrule

        Residual PMA + Alloc.
        & 26.9088
        & 0.8550
        & 0.0791
        \\

        \addlinespace[2pt]

        Direction Only
        & 26.8438
        & 0.8507
        & 0.0899
        \\

        Moment Only
        & 27.2093
        & 0.8600
        & 0.0832
        \\

        \addlinespace[2pt]

        w/o Top-1 Preservation
        & 23.1402
        & 0.7202
        & 0.1894
        \\

        w/o Entropy Floor
        & 24.7703
        & 0.7804
        & 0.1060
        \\

        \midrule

        \textbf{Full TACC}
        & \textbf{27.3415}
        & \textbf{0.8649}
        & \textbf{0.0791}
        \\

        \bottomrule

    \end{tabular}
    }
\end{table}

As shown in Tab.~\ref{tab:controlled_ablations}, under the shared
allocation rule, Residual PMA + Alloc. achieves $26.91$~dB PSNR,
compared with $27.34$~dB for TACC, with the same reported LPIPS
in this setting.

As shown in Tab.~\ref{tab:allocation_ablations}, moment distance yields
higher reconstruction quality than directional similarity alone in this
setting. Combining the two terms further improves PSNR from $27.21$ to
$27.34$~dB, with corresponding improvements in SSIM and LPIPS.

Removing Top-1 preservation reduces PSNR from 27.34 to
23.14 dB, while removing the entropy-based coverage
floor yields 24.77 dB. The respective reductions of
4.20 and 2.57 dB indicate that both safeguards contribute
substantially to reconstruction quality.

\begingroup

\paragraph{Training-loss ablations.}
We further assess the contributions of the distillation
and Carrier reconstruction losses defined in
Appendix~\ref{app:carrier_training}.
Unlike the inference-time allocation ablations above,
each loss ablation independently retrains the Carrier
with the corresponding loss removed, while keeping
the LagerNVS backbone frozen and the remaining
training configuration unchanged.

We evaluate the resulting models on the official
140-scene DL3DV benchmark with 80, 100, and
120 source views, a fixed $A=10$ decoder-visible
budget, and 20 target views per scene.
As reported in Tab.~\ref{tab:app_training_loss_ablations},
removing either loss consistently degrades reconstruction
quality. Averaged over the three source-view settings,
PSNR decreases by 0.56 dB without the distillation
loss and by 0.62 dB without the Carrier loss.

\par
\endgroup

\begin{table}[t]
    \centering
    {%
    \caption{
        \textbf{Training-loss ablations on DL3DV.}
        LagerNVS is evaluated with a fixed $A=10$
        budget. Each loss-ablated Carrier is independently
        retrained with the corresponding objective removed.
        The backbone remains frozen.
        Eff. FPS denotes Effective FPS.
    }
    \label{tab:app_training_loss_ablations}

    \small
    \setlength{\tabcolsep}{4.0pt}
    \renewcommand{\arraystretch}{1.08}

    \begin{tabular}{@{}clcccc@{}}
        \toprule

        \textbf{Views}
        & \textbf{Variant}
        & \textbf{PSNR} $\uparrow$
        & \textbf{SSIM} $\uparrow$
        & \textbf{LPIPS} $\downarrow$
        & \textbf{Eff. FPS} $\uparrow$
        \\

        \midrule

        80
        & \textbf{Full TACC}
        & \textbf{27.3415}
        & \textbf{0.8649}
        & \textbf{0.0791}
        & 27.95
        \\

        80
        & w/o Distillation Loss
        & 26.6974
        & 0.8517
        & 0.0816
        & 28.11
        \\

        80
        & w/o Carrier Loss
        & 26.5642
        & 0.8480
        & 0.0825
        & 28.16
        \\

        \midrule

        100
        & \textbf{Full TACC}
        & \textbf{26.8787}
        & \textbf{0.8523}
        & \textbf{0.0833}
        & 28.12
        \\

        100
        & w/o Distillation Loss
        & 26.3766
        & 0.8417
        & 0.0848
        & 28.10
        \\

        100
        & w/o Carrier Loss
        & 26.2987
        & 0.8400
        & 0.0851
        & 28.07
        \\

        \midrule

        120
        & \textbf{Full TACC}
        & \textbf{26.4732}
        & \textbf{0.8416}
        & \textbf{0.0869}
        & 27.96
        \\

        120
        & w/o Distillation Loss
        & 25.9353
        & 0.8305
        & 0.0884
        & 27.81
        \\

        120
        & w/o Carrier Loss
        & 25.9747
        & 0.8320
        & 0.0879
        & 28.06
        \\

        \bottomrule

    \end{tabular}
    }
\end{table}


\begingroup

\paragraph{Geometry-conditioning-inspired compression adapters.}
To compare alternative designs under a common compression setting,
we implement three adapters inspired by the projected representation
conditioning of ReNoV~\citep{kwak2026renov}, the projective
conditioning of PVSM~\citep{wu2026projections}, and the
confidence-aware visual geometry routing of
VGGT-Diff~\citep{chen2026vggtdiff}.
These adapters are evaluated as fixed-budget source-representation
compressors within the frozen LagerNVS rendering pipeline,
rather than as reproductions of the original complete NVS systems.

Specifically, each adapter operates between the LagerNVS Reconstructor
and Renderer, replacing their intermediate source-token representation
with exactly $B=13{,}690$ tokens ($A=10$): $117\times117$
target-aligned spatial tokens and one pooled source token.
The pretrained Reconstructor and Renderer remain frozen, and only
the respective adapter is trained.
The three adapters instantiate this compression interface as follows.

\emph{ReNoV-inspired.}
We associate source features with 3D positions recovered from source
depth, project them onto the target-view grid, and combine the projected
features with Fourier-encoded coordinates, camera information, and
visibility masks to construct the compressed representation.

\emph{PVSM-inspired.}
We construct colored 3D points from source RGB, depth, and camera poses.
Gaussian splatting produces target-view RGB and alpha maps, which a
lightweight encoder maps into spatial tokens; a pooled source token
completes the fixed-budget representation.

\emph{VGR-inspired.}
We combine source features with back-projected 3D positions and
confidence estimates from a frozen VGGT model. Depth- and
confidence-guided projection produces a near-surface hard anchor,
which is refined by residual fusion of front- and back-surface evidence.

Unlike TACC, these adapters do not use progressive Carrier hierarchies
or target-conditioned per-view quota allocation; instead, they compress
source information through their respective geometric projection and
aggregation mechanisms.

All three adapted baselines follow the same initialization protocol,
training schedule, and optimization settings as TACC, using the
same frozen LagerNVS backbone and DL3DV training data.
We evaluate all methods on the official 140-scene DL3DV benchmark
with identical source and target views, 20 target views per scene,
80, 100, and 120 source views, and a fixed $A=10$
decoder-visible token budget.
The results are reported in Tab.~\ref{tab:geometry_adapters}.

\par

\begin{table}[t]
    \centering
    \caption{
        \textbf{Comparison with geometry-conditioning-inspired
        compression adapters on DL3DV.}
        All methods use LagerNVS and a fixed $A=10$
        decoder-visible budget.
        Params denotes the number of trainable parameters.
        Reconstruction quality is reported in PSNR (dB).
    }
    \label{tab:geometry_adapters}

    \small
    \setlength{\tabcolsep}{4.5pt}
    \renewcommand{\arraystretch}{1.08}

    \begin{tabular}{@{}lcccc@{}}
        \toprule

        \textbf{Method}
        & \textbf{Params}
        & \textbf{80v} $\uparrow$
        & \textbf{100v} $\uparrow$
        & \textbf{120v} $\uparrow$
        \\

        \midrule

        ReNoV-inspired
        & 0.454M
        & 23.0151
        & 22.7607
        & 22.4773
        \\

        PVSM-inspired
        & 0.155M
        & 13.5317
        & 13.5265
        & 13.5201
        \\

        VGR-inspired
        & 0.413M
        & 22.6085
        & 22.3263
        & 22.0373
        \\

        \midrule

        \textbf{TACC}
        & \textbf{0.303M}
        & \textbf{27.3415}
        & \textbf{26.8787}
        & \textbf{26.4732}
        \\

        \bottomrule

    \end{tabular}
\end{table}

Across all three source-view settings, TACC outperforms
the strongest adapted baseline by 4.00--4.33 dB in PSNR.
These results support the effectiveness of TACC's progressive
Carrier representation and target-conditioned allocation
under a fixed decoder-visible token budget.

\par
\endgroup

\subsection{Target-Time and End-to-End Runtime Analysis}
\label{app:runtime}

\paragraph{Target-time profiling.}
We measure target-time latency on an NVIDIA A800 GPU under BF16
inference. One complete scene is used for warm-up and excluded from
measurement. GPU execution is synchronized at the boundaries of each
timed region. Stage-wise profiling and end-to-end timing are performed
in separate runs to avoid introducing profiling synchronization into
the end-to-end measurements.

For the target-time analysis, we use the official 140-scene DL3DV
evaluation set and the Front180 protocol described in
Appendix~\ref{app:view_selection}, with 20 target views per scene
and a target-view batch size of 4. We profile allocation, frontier
readout, and decoder execution separately. For each stage, we
accumulate latency across all target batches of a scene, average
over the 140 scenes, and divide by the 20 rendered target views.
Effective FPS is computed from the corresponding target-time latency.

As shown in Tab.~\ref{tab:main_runtime_costs}
in Sec.~4.2, the fixed decoder-visible budget keeps TACC's
decoder latency nearly unchanged as the source context grows
from 80 to 120 views, whereas Full Context becomes progressively
slower. Allocation and frontier readout introduce a small
per-target overhead. The decoder speedups are $6.94\times$,
$8.62\times$, and $10.31\times$ at 80, 100, and 120 source views,
respectively, with corresponding Effective FPS of 27.95, 28.12,
and 27.96.

\paragraph{Amortized end-to-end throughput.}
Effective FPS excludes source encoding and the one-time Carrier
construction. To examine the amortized cost of these operations,
we evaluate LagerNVS on the 140-scene DL3DV benchmark and
LVSM Decoder-Only on the 1,633-scene RE10K evaluation set.
In the original target-count sweeps,
both use 80 source views and a fixed $A=10$ decoder-visible
budget for TACC, while varying the number of rendered targets.

Both backbones follow the same target-list construction procedure,
using the first 180 valid frames for LagerNVS on DL3DV and the
first 200 valid frames for LVSM Decoder-Only on RE10K, as defined
in Appendix~\ref{app:view_selection}. For each scene, we begin
with the stored 40-view target list in ascending frame-index order.
We then append 60 additional targets from the remaining candidate
frames, excluding the fixed 80 source views and previously selected
targets. These additional views are selected sequentially using
deterministic farthest-first selection along the cumulative pose
coordinate defined in Appendix~\ref{app:view_selection}.

\begin{figure}[H]
    \vspace{-3pt}
    \centering
    \includegraphics[
        width=0.50\linewidth
    ]{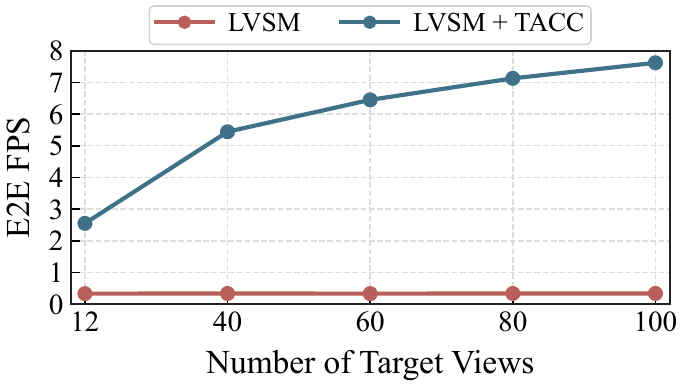}

    \vspace{-7pt}

    \caption{
        \textbf{Amortized E2E throughput on LVSM Decoder-Only.}
        E2E FPS on RE10K with 80 source views and a fixed
        $A=10$ decoder-visible budget for TACC.
    }
    \label{fig:amortized_e2e}
\end{figure}

For LagerNVS, we evaluate target counts
$K\in\{10,40,60,80,100\}$; for LVSM Decoder-Only, we use
$K\in\{12,40,60,80,100\}$. Each target set contains the first
$K$ entries of the corresponding ordered 100-view list.
The target sets are therefore nested and disjoint from the fixed
source views. Full Context and TACC use identical source and
target lists at every target count. TACC computes relevance scores
and per-view quotas independently for each target; the target-view
batch size of 4 affects decoder batching but not target selection.

We measure E2E runtime from source-view processing through target
RGB prediction, including the complete model computation but
excluding disk loading and image decoding. Source encoding is
performed once per source set for both methods, and TACC
additionally constructs the Carrier hierarchies once. Each of
the 140 DL3DV scenes and 1,633 RE10K scenes is measured once
at every target count under the corresponding backbone.
E2E FPS is computed as the total number of rendered targets
divided by the total measured wall-clock time.

The LagerNVS E2E results are presented in
Fig.~\ref{fig:lvsm_efficiency}(c).
The complementary LVSM Decoder-Only results in
Fig.~\ref{fig:amortized_e2e} show higher E2E throughput for TACC
across all evaluated target counts.

\paragraph{E2E near-parity across source-view counts.}
We further evaluate LagerNVS on the official 140-scene
DL3DV benchmark with source-view counts
$V\in\{15,20,40,60,80,100,120\}$ and a fixed
$A=10$ decoder-visible budget. We follow the same E2E timing
protocol, including source encoding, one-time Carrier
construction, and target rendering, with a target-view batch size
of 4. Source-view sets follow the nested sampling protocol in
Appendix~\ref{app:view_selection}. Unlike the original fixed-80-view
target-count sweeps, target views in this experiment are evenly
sampled from the remaining non-source frames for each source-view
setting. Full Context and TACC use identical source and target
views within each setting.

For each source-view count, we evaluate multiple
target counts and select the measured point with the smallest
absolute E2E FPS difference between Full Context and TACC.
The corresponding results are listed in
Tab.~\ref{tab:app_e2e_near_parity} and visualized in
Fig.~\ref{fig:tacc_quality_efficiency}(b).
These points represent approximate E2E throughput parity,
not exact acceleration thresholds.

\begin{table}[H]
    \centering
    {%
    \caption{
        \textbf{Measured E2E near-parity points on DL3DV.}
        LagerNVS uses a fixed $A=10$ budget.
        E2E FPS includes one-time source encoding
        and Carrier construction.
    }
    \label{tab:app_e2e_near_parity}

    \small
    \setlength{\tabcolsep}{7pt}
    \renewcommand{\arraystretch}{1.08}

    \begin{tabular}{@{}cccc@{}}
        \toprule
        \textbf{Source Views}
        & \textbf{Target Views}
        & \textbf{Full E2E FPS}
        & \textbf{TACC E2E FPS} \\
        \midrule
        15  & 48 & 16.4181 & 16.1030 \\
        20  & 32 & 10.9359 & 10.9403 \\
        40  & 20 & 3.8171  & 3.8998 \\
        60  & 17 & 1.9860  & 1.9939 \\
        80  & 16 & 1.2514  & 1.2523 \\
        100 & 16 & 0.8928  & 0.8839 \\
        120 & 16 & 0.6722  & 0.6636 \\
        \bottomrule
    \end{tabular}
    }
\end{table}

As shown in Tab.~\ref{tab:app_e2e_near_parity},
the near-parity target count decreases from 48 at
15 source views to 16 at 80--120 source views.
At the reported target counts, TACC achieves slightly
higher E2E FPS at 20, 40, 60, and 80 source views,
but remains slightly slower at 15, 100, and
120 source views. These measurements illustrate
the amortization trend without establishing
exact break-even thresholds.%

%
We additionally evaluate LagerNVS with 80, 100, and 120 source
views using finer target counts near E2E parity, following the
same E2E timing protocol and a target-view batch size of 4.
Tab.~\ref{tab:main_runtime_costs} in Sec.~4.2 reports the
Carrier construction time and reusable source-cache sizes,
while Tab.~\ref{tab:app_e2e_acceleration_points} summarizes
the measured near-parity and acceleration target counts.
Construction time is averaged over the 140 DL3DV scenes and
includes hierarchy construction, frontier-table precomputation,
and Carrier network precomputation. Cache sizes include all
retained source-side representations on the GPU, with shared
storage counted once, excluding model weights and temporary
target-side tensors.%

\begin{table}[H]
    \centering
    {
    \caption{
        \textbf{Measured E2E near-parity and acceleration
        points on DL3DV.}
        Results use LagerNVS with a fixed $A=10$ budget.
        Target counts are measured reference points rather
        than exact break-even thresholds.
    }
    \label{tab:app_e2e_acceleration_points}

    \small
    \setlength{\tabcolsep}{8pt}
    \renewcommand{\arraystretch}{1.08}

    \begin{tabular}{@{}ccc@{}}
        \toprule
        \textbf{Source Views}
        & \textbf{Near-Parity Targets}
        & \textbf{Measured Acceleration Targets} \\
        \midrule
        80  & 16 & 20 \\
        100 & 16 & 18 \\
        120 & 16 & 18 \\
        \bottomrule
    \end{tabular}
    }
\end{table}

At 16 target views, TACC's total runtime differs from Full Context
by $-0.08\%$, $+1.01\%$, and $+1.30\%$ for 80, 100, and 120
source views, respectively. We therefore treat 16 targets as a
near-parity reference rather than an exact break-even threshold.
Measured E2E acceleration occurs at 20 targets for 80 source views
and at 18 targets for 100 and 120 source views.
At 80 source views and 20 targets, TACC improves E2E FPS
by approximately $7.3\%$ (Sec.~4.2).%

TACC retains additional source-side representations, including
regional means, hierarchical Carriers, and source geometry,
resulting in a larger reusable cache than Full Context.
These cache sizes should not be interpreted as overall peak
GPU memory. The fixed decoder-visible budget limits the
source-token context processed during rendering, while the
reusable cache incurs a separate storage cost.%

\subsection{Robustness of Geometric Relevance and Allocation}
\label{app:translation_robustness}

For LagerNVS, the distance term in Eq.~(5) uses the Pl\"ucker moment
$m=o\times d$ of the center ray. This quantity depends on the coordinate
origin: under a common translation $u$ applied to all camera origins,
$o'=o+u$, the moment becomes $m'=m+u\times d$, while $d$ remains unchanged.
We therefore evaluate the sensitivity of the LagerNVS allocation rule to
common translations applied only to the Pl\"ucker rays used for relevance
computation.

\begin{table}[!htbp]
    \centering
    \caption{
        \textbf{Sensitivity of LagerNVS allocation to common translations of
        the allocation coordinates.}
        Translation labels denote scene-radius-scaled per-axis coefficients.
        Top-1 Change measures changes in the highest-relevance source view,
        while Quota TV measures the normalized total variation of the
        resulting per-view allocation.
    }
    \label{tab:translation_robustness}

    \small
    \setlength{\tabcolsep}{5.0pt}
    \renewcommand{\arraystretch}{1.08}

    \resizebox{\linewidth}{!}{
    \begin{tabular}{@{}lccccc@{}}
        \toprule

        \textbf{Translation}
        & \textbf{PSNR} $\uparrow$
        & \textbf{SSIM} $\uparrow$
        & \textbf{LPIPS} $\downarrow$
        & \textbf{Top-1 Change} $\downarrow$
        & \textbf{Quota TV} $\downarrow$
        \\

        \midrule

        None
        & 27.5815
        & 0.8691
        & 0.0792
        & --
        & --
        \\

        $x=0.1$
        & 27.5854
        & 0.8692
        & 0.0792
        & 1.04\%
        & 0.292\%
        \\

        $x=0.5$
        & 27.5722
        & 0.8692
        & 0.0792
        & 4.64\%
        & 0.700\%
        \\

        $x=1.0$
        & 27.5505
        & 0.8685
        & 0.0796
        & 7.96\%
        & 1.052\%
        \\

        $y=0.5$
        & 27.5829
        & 0.8692
        & 0.0793
        & 4.04\%
        & 0.602\%
        \\

        $z=0.5$
        & 27.5678
        & 0.8686
        & 0.0797
        & 3.86\%
        & 0.554\%
        \\

        $xyz=0.5$
        & 27.5585
        & 0.8686
        & 0.0797
        & 6.07\%
        & 0.807\%
        \\

        \bottomrule
    \end{tabular}
    }
\end{table}

We use LagerNVS on DL3DV with 60 source views, 20 target views, and an
$A=10$ decoder-visible budget. The perturbation is applied after the standard
LagerNVS camera normalization, which recenters poses with respect to the first
camera and scales camera translations by $1.35$ times the maximum
source-camera translation norm. For each scene, let $\rho$ denote the maximum
distance of the 60 source and 20 target camera centers from their joint
centroid in this normalized coordinate system. We apply
$u=\rho\gamma$, where the table labels specify the per-axis coefficient
$\gamma$; for example, $x=0.5$ denotes $\gamma=(0.5,0,0)$, while
$xyz=0.5$ denotes $\gamma=(0.5,0.5,0.5)$.
The same $u$ is applied to all source and target Pl\"ucker rays used for
allocation within a scene. Source and target view indices, rendering inputs,
evaluation images, and model parameters remain unchanged.

Let $q_{i,t}^{(0)}$ and $q_{i,t}^{(u)}$ denote the quota assigned to source
view $i$ for target view $t$ before and after translation, respectively, with
$\sum_i q_{i,t}^{(0)}=\sum_i q_{i,t}^{(u)}=B$. We measure the allocation
change for each target view as
\begin{equation}
\operatorname{TV}_t
=
\frac{1}{2B}
\sum_{i=1}^{V}
\left|
q_{i,t}^{(u)}-q_{i,t}^{(0)}
\right|,
\qquad
\operatorname{Top1}_t
=
\mathbf{1}\!\left[
\arg\max_i r_{i,t}^{(u)}
\neq
\arg\max_i r_{i,t}^{(0)}
\right].
\label{eq:app_translation_metrics}
\end{equation}
Quota TV and Top-1 Change are averaged over all
$140\times20=2{,}800$ scene--target-view pairs and reported as percentages.

As shown in Tab.~\ref{tab:translation_robustness}, common translations
produce modest changes in the resulting allocation.
The largest mean Quota TV is $1.052\%$, attained for $x=1.0$, where the
highest-relevance source view changes for $7.96\%$ of the evaluated targets.
The corresponding reconstruction metrics vary little, with a maximum PSNR
decrease of approximately $0.031$~dB and similarly small changes in SSIM and
LPIPS. Although the LagerNVS moment-distance term is not strictly translation
invariant, its allocation remains empirically stable under the tested
coordinate perturbations.

\noindent\textbf{Source-pose perturbations and outliers.}
We further evaluate LagerNVS on the official 140-scene DL3DV benchmark
with 80 source views, 20 target views per scene, and a fixed $A=10$
budget. To isolate the allocation mechanism, we perturb only the
source-camera poses used for relevance computation; encoded features,
Carrier hierarchies, target rendering rays, and model parameters
remain unchanged.

We consider three perturbation magnitudes applied
independently to all 80 source cameras, using rotations of
$1^\circ$, $3^\circ$, and $5^\circ$ with corresponding translations
of $1\%R$, $3\%R$, and $5\%R$, respectively. We also evaluate sparse
outliers affecting four randomly selected source cameras, using
$10^\circ$ rotations and translations of $10\%R$. Here, $R$ is the
root-mean-square distance of source-camera centers from their
centroid. Rotation axes and translation directions are sampled
independently at random. We evaluate three fixed random seeds,
each covering 140 scenes and 2,800 target views.

\begin{table}[!htbp]
    \centering
    {%
    \caption{
        \textbf{Sensitivity to source-camera pose perturbations on DL3DV.}
        Only source poses used for allocation are perturbed.
        LagerNVS uses 80 source views and $A=10$.
        PSNR is averaged over three random seeds;
        $\Delta$PSNR reports the mean change relative to original poses
        and its standard deviation across the three seeds.
    }
    \label{tab:app_pose_perturbation}

    \small
    \setlength{\tabcolsep}{5pt}
    \renewcommand{\arraystretch}{1.08}

    \resizebox{\linewidth}{!}{%
    \begin{tabular}{@{}lcccc@{}}
        \toprule
        \textbf{Setting}
        & \textbf{Cameras}
        & \textbf{Rotation / Translation}
        & \textbf{PSNR} $\uparrow$
        & \shortstack{\textbf{$\Delta$PSNR (dB)}\\
                      \textbf{mean $\pm$ std}}
        \\
        \midrule

        Original Poses
        & 0/80
        & --
        & 27.3415
        & --
        \\

        Small Perturbations
        & 80/80
        & $1^\circ$ / $1\%R$
        & 27.2625
        & $-0.0790 \pm 0.0113$
        \\

        Moderate Perturbations
        & 80/80
        & $3^\circ$ / $3\%R$
        & 26.9075
        & $-0.4340 \pm 0.0256$
        \\

        Large Perturbations
        & 80/80
        & $5^\circ$ / $5\%R$
        & 26.4560
        & $-0.8855 \pm 0.0246$
        \\

        Sparse Outliers
        & 4/80
        & $10^\circ$ / $10\%R$
        & 27.3048
        & $-0.0367 \pm 0.0065$
        \\

        \bottomrule
    \end{tabular}%
    }
    }
\end{table}

As shown in Tab.~\ref{tab:app_pose_perturbation},
the PSNR decreases by $0.079$, $0.434$, and $0.886$~dB under
$1^\circ$, $3^\circ$, and $5^\circ$ perturbations, respectively,
indicating increasing sensitivity as the perturbation magnitude
grows. Sparse outliers affecting four source cameras cause only
a $0.037$~dB decrease. These results characterize the allocation
mechanism's sensitivity to the tested source-pose errors.
They do not establish robustness of the full rendering pipeline
to erroneous camera poses, because the encoded source
representations and rendering rays are unchanged.

\noindent\textbf{Uniform allocation under similar relevance.}
We compare TACC with an inference-time Uniform Allocation control
that distributes the same $A=10$ token budget approximately equally
across all 80 source views. Both methods use identical source and
target views and the same trained Carrier checkpoint; unlike the
retrained Uniform Alloc. ablation in Sec.~4.3, no new Carrier is
trained for this control.

For each of the 2,800 target views, we compute the
difference between the first- and tenth-ranked source-view
relevance scores. The 25\% with the smallest differences form
the similar-relevance subset (700 targets); the remaining 2,100
targets form the complementary subset.

\begin{table}[!htbp]
    \centering
    {%
    \caption{
        \textbf{Uniform allocation under similar relevance scores on DL3DV.}
        Both methods reuse the same Carrier checkpoint at $A=10$.
        PSNR gain denotes TACC minus Uniform Allocation.
    }
    \label{tab:app_near_uniform}
    \small
    \setlength{\tabcolsep}{5pt}
    \renewcommand{\arraystretch}{1.08}
    \begin{tabular}{@{}lcccc@{}}
        \toprule
        \textbf{Target Group} & \textbf{Targets}
        & \shortstack{\textbf{PSNR Gain}\\\textbf{(dB)}}
        & \shortstack{\textbf{Uniform}\\\textbf{Wins}}
        & \shortstack{\textbf{Win}\\\textbf{Rate}} \\
        \midrule
        Smallest 25\% Gaps & 700   & +4.9928 & 25 & 3.57\% \\
        Remaining 75\%     & 2,100 & +4.1131 & 72 & 3.43\% \\
        \bottomrule
    \end{tabular}
    }
\end{table}

As shown in Tab.~\ref{tab:app_near_uniform},
Uniform Allocation outperforms TACC on 25 of the 700 similar-relevance
targets (3.57\%) and 72 of the remaining 2,100 targets (3.43\%).
TACC nevertheless achieves mean PSNR advantages of 4.99 and 4.11~dB,
respectively. Thus, small relevance-score gaps do not systematically
favor uniform allocation. Such gaps quantify score separation,
not the correctness of the underlying geometric relevance estimates;
systematic relevance-estimation errors are not assessed here.

%% file: iclr2027_conference.bib
@inproceedings{pixelsplat,
  author    = {Charatan, David and Li, Sizhe Lester and
               Tagliasacchi, Andrea and Sitzmann, Vincent},
  title     = {{pixelSplat}: {3D} Gaussian Splats from Image Pairs for
               Scalable Generalizable {3D} Reconstruction},
  booktitle = {Proceedings of the IEEE/CVF Conference on Computer Vision
               and Pattern Recognition (CVPR)},
  pages     = {19457--19467},
  month     = jun,
  year      = {2024}
}

@inproceedings{nerf,
  author    = {Mildenhall, Ben and Srinivasan, Pratul P. and Tancik, Matthew and
               Barron, Jonathan T. and Ramamoorthi, Ravi and Ng, Ren},
  title     = {{NeRF}: Representing Scenes as Neural Radiance Fields for
               View Synthesis},
  booktitle = {Computer Vision -- ECCV 2020},
  pages     = {405--421},
  publisher = {Springer},
  year      = {2020}
}

@article{3dgs,
  author  = {Kerbl, Bernhard and Kopanas, Georgios and
             Leimk{\"u}hler, Thomas and Drettakis, George},
  title   = {{3D} Gaussian Splatting for Real-Time Radiance Field Rendering},
  journal = {ACM Transactions on Graphics},
  volume  = {42},
  number  = {4},
  pages   = {139:1--139:14},
  month   = jul,
  year    = {2023}
}

@inproceedings{mvsplat,
  author    = {Chen, Yuedong and Xu, Haofei and Zheng, Chuanxia and
               Zhuang, Bohan and Pollefeys, Marc and Geiger, Andreas and
               Cham, Tat-Jen and Cai, Jianfei},
  title     = {{MVSplat}: Efficient {3D} Gaussian Splatting from Sparse
               Multi-View Images},
  booktitle = {Computer Vision -- ECCV 2024},
  pages     = {370--386},
  year      = {2024}
}

@inproceedings{srt,
  author    = {Sajjadi, Mehdi S. M. and Meyer, Henning and Pot, Etienne and
               Bergmann, Urs and Greff, Klaus and Radwan, Noha and
               Vora, Suhani and Lu\v{c}i\'{c}, Mario and Duckworth, Daniel and
               Dosovitskiy, Alexey and Uszkoreit, Jakob and
               Funkhouser, Thomas and Tagliasacchi, Andrea},
  title     = {Scene Representation Transformer: Geometry-Free Novel View
               Synthesis Through Set-Latent Scene Representations},
  booktitle = {Proceedings of the IEEE/CVF Conference on Computer Vision
               and Pattern Recognition (CVPR)},
  pages     = {6229--6238},
  month     = jun,
  year      = {2022}
}

@article{hinton2015distilling,
  author  = {Hinton, Geoffrey and Vinyals, Oriol and Dean, Jeff},
  title   = {Distilling the Knowledge in a Neural Network},
  journal = {arXiv preprint arXiv:1503.02531},
  year    = {2015}
}

@inproceedings{lvsm,
  author    = {Jin, Haian and Jiang, Hanwen and Tan, Hao and Zhang, Kai and
               Bi, Sai and Zhang, Tianyuan and Luan, Fujun and
               Snavely, Noah and Xu, Zexiang},
  title     = {{LVSM}: A Large View Synthesis Model with Minimal {3D}
               Inductive Bias},
  booktitle = {The Thirteenth International Conference on Learning
               Representations},
  year      = {2025}
}

@inproceedings{vggt,
  author    = {Wang, Jianyuan and Chen, Minghao and Karaev, Nikita and
               Vedaldi, Andrea and Rupprecht, Christian and Novotny, David},
  title     = {{VGGT}: Visual Geometry Grounded Transformer},
  booktitle = {Proceedings of the IEEE/CVF Conference on Computer Vision
               and Pattern Recognition (CVPR)},
  pages     = {5294--5306},
  month     = jun,
  year      = {2025}
}

@inproceedings{lagernvs,
  author    = {Szymanowicz, Stanislaw and Chen, Minghao and
               Wang, Jianyuan and Rupprecht, Christian and Vedaldi, Andrea},
  title     = {{LagerNVS}: Latent Geometry for Fully Neural Real-time
               Novel View Synthesis},
  booktitle = {Proceedings of the IEEE/CVF Conference on Computer Vision
               and Pattern Recognition (CVPR)},
  pages     = {15443--15453},
  month     = jun,
  year      = {2026}
}

@inproceedings{dynamicvit,
  author    = {Rao, Yongming and Zhao, Wenliang and Liu, Benlin and
               Lu, Jiwen and Zhou, Jie and Hsieh, Cho-Jui},
  title     = {{DynamicViT}: Efficient Vision Transformers with Dynamic
               Token Sparsification},
  booktitle = {Advances in Neural Information Processing Systems},
  volume    = {34},
  pages     = {13937--13949},
  year      = {2021}
}

@inproceedings{evit,
  author    = {Liang, Youwei and Ge, Chongjian and Tong, Zhan and
               Song, Yibing and Wang, Jue and Xie, Pengtao},
  title     = {Not All Patches Are What You Need: Expediting Vision
               Transformers via Token Reorganizations},
  booktitle = {International Conference on Learning Representations},
  year      = {2022}
}

@inproceedings{tokenlearner,
  author    = {Ryoo, Michael S. and Piergiovanni, A. J. and Arnab, Anurag and
               Dehghani, Mostafa and Angelova, Anelia},
  title     = {{TokenLearner}: Adaptive Space-Time Tokenization for Videos},
  booktitle = {Advances in Neural Information Processing Systems},
  volume    = {34},
  pages     = {12786--12797},
  year      = {2021}
}

@inproceedings{tome,
  author    = {Bolya, Daniel and Fu, Cheng-Yang and Dai, Xiaoliang and
               Zhang, Peizhao and Feichtenhofer, Christoph and Hoffman, Judy},
  title     = {Token Merging: Your {ViT} But Faster},
  booktitle = {The Eleventh International Conference on Learning
               Representations},
  year      = {2023}
}

@inproceedings{divprune,
  author    = {Alvar, Saeed Ranjbar and Singh, Gursimran and
               Akbari, Mohammad and Zhang, Yong},
  title     = {{DivPrune}: Diversity-based Visual Token Pruning for
               Large Multimodal Models},
  booktitle = {Proceedings of the IEEE/CVF Conference on Computer Vision
               and Pattern Recognition (CVPR)},
  pages     = {9392--9401},
  month     = jun,
  year      = {2025}
}

@inproceedings{fastvggt,
  author    = {Shen, You and Zhang, Zhipeng and Qu, Yansong and
               Zheng, Xiawu and Ji, Jiayi and Zhang, Shengchuan and
               Cao, Liujuan},
  title     = {{FastVGGT}: Fast Visual Geometry Transformer},
  booktitle = {The Fourteenth International Conference on Learning
               Representations},
  year      = {2026}
}

@article{good_token_hunting,
  author  = {Zheng, Shuhong and Oechsle, Michael and
             Sandstr{\"o}m, Erik and Rakotosaona, Marie-Julie and
             Tombari, Federico and Gilitschenski, Igor},
  title   = {Good Token Hunting: A Hitchhiker's Guide to Token Selection
             for Visual Geometry Transformers},
  journal = {arXiv preprint arXiv:2605.23892},
  year    = {2026}
}

@inproceedings{come,
  author    = {Chen, Yutian and Qiu, Yuheng and Li, Ruogu and
               Patrikar, Jay and Scherer, Sebastian},
  title     = {{Co-Me}: Confidence Guided Token Merging for Visual
               Geometric Transformers},
  booktitle = {Proceedings of the IEEE/CVF Conference on Computer Vision
               and Pattern Recognition (CVPR)},
  pages     = {14590--14599},
  month     = jun,
  year      = {2026}
}

@inproceedings{zpressor,
  author    = {Wang, Weijie and Chen, Donny Y. and Zhang, Zeyu and
               Shi, Duochao and Liu, Akide and Zhuang, Bohan},
  title     = {{ZPressor}: Bottleneck-Aware Compression for Scalable
               Feed-Forward {3DGS}},
  booktitle = {Advances in Neural Information Processing Systems},
  volume    = {38},
  pages     = {125793--125822},
  year      = {2025}
}

@inproceedings{clift,
  author    = {Wang, Zhengqing and Wu, Yuefan and Chen, Jiacheng and
               Zhang, Fuyang and Furukawa, Yasutaka},
  title     = {{CLiFT}: Compressive Light-Field Tokens for Compute
               Efficient and Adaptive Neural Rendering},
  booktitle = {Advances in Neural Information Processing Systems},
  volume    = {38},
  pages     = {8759--8787},
  year      = {2025}
}

@inproceedings{dl3dv,
  author    = {Ling, Lu and Sheng, Yichen and Tu, Zhi and
               Zhao, Wentian and Xin, Cheng and Wan, Kun and
               Yu, Lantao and Guo, Qianyu and Yu, Zixun and
               Lu, Yawen and Li, Xuanmao and Sun, Xingpeng and
               Ashok, Rohan and Mukherjee, Aniruddha and
               Kang, Hao and Kong, Xiangrui and Hua, Gang and
               Zhang, Tianyi and Benes, Bedrich and Bera, Aniket},
  title     = {{DL3DV-10K}: A Large-Scale Scene Dataset for
               Deep Learning-based {3D} Vision},
  booktitle = {Proceedings of the IEEE/CVF Conference on Computer Vision
               and Pattern Recognition (CVPR)},
  pages     = {22160--22169},
  month     = jun,
  year      = {2024}
}

@article{realestate10k,
  author  = {Zhou, Tinghui and Tucker, Richard and Flynn, John and
             Fyffe, Graham and Snavely, Noah},
  title   = {Stereo Magnification: Learning View Synthesis Using
             Multiplane Images},
  journal = {ACM Transactions on Graphics},
  volume  = {37},
  number  = {4},
  pages   = {65:1--65:12},
  month   = aug,
  year    = {2018}
}

@article{ssim,
  author  = {Wang, Zhou and Bovik, Alan C. and Sheikh, Hamid R. and
             Simoncelli, Eero P.},
  title   = {Image Quality Assessment: From Error Visibility to
             Structural Similarity},
  journal = {IEEE Transactions on Image Processing},
  volume  = {13},
  number  = {4},
  pages   = {600--612},
  month   = apr,
  year    = {2004}
}

@inproceedings{lpips,
  author    = {Zhang, Richard and Isola, Phillip and Efros, Alexei A. and
               Shechtman, Eli and Wang, Oliver},
  title     = {The Unreasonable Effectiveness of Deep Features as a
               Perceptual Metric},
  booktitle = {Proceedings of the IEEE Conference on Computer Vision
               and Pattern Recognition (CVPR)},
  pages     = {586--595},
  month     = jun,
  year      = {2018}
}

@inproceedings{ibrnet,
  author    = {Wang, Qianqian and Wang, Zhicheng and Genova, Kyle and
               Srinivasan, Pratul P. and Zhou, Howard and
               Barron, Jonathan T. and Martin-Brualla, Ricardo and
               Snavely, Noah and Funkhouser, Thomas},
  title     = {{IBRNet}: Learning Multi-View Image-Based Rendering},
  booktitle = {Proceedings of the IEEE/CVF Conference on Computer Vision
               and Pattern Recognition (CVPR)},
  pages     = {4690--4699},
  month     = jun,
  year      = {2021}
}

@inproceedings{mvsnerf,
  author    = {Chen, Anpei and Xu, Zexiang and Zhao, Fuqiang and
               Zhang, Xiaoshuai and Xiang, Fanbo and Yu, Jingyi and Su, Hao},
  title     = {{MVSNeRF}: Fast Generalizable Radiance Field Reconstruction
               from Multi-View Stereo},
  booktitle = {Proceedings of the IEEE/CVF International Conference on
               Computer Vision (ICCV)},
  pages     = {14124--14133},
  month     = oct,
  year      = {2021}
}

@inproceedings{geonerf,
  author    = {Johari, Mohammad Mahdi and Lepoittevin, Yann and
               Fleuret, Fran\c{c}ois},
  title     = {{GeoNeRF}: Generalizing {NeRF} with Geometry Priors},
  booktitle = {Proceedings of the IEEE/CVF Conference on Computer Vision
               and Pattern Recognition (CVPR)},
  pages     = {18365--18375},
  month     = jun,
  year      = {2022}
}

@inproceedings{viewformer,
  author    = {Kulh{\'a}nek, Jon{\'a}{\v{s}} and Derner, Erik and
               Sattler, Torsten and Babu{\v{s}}ka, Robert},
  title     = {{ViewFormer}: {NeRF}-Free Neural Rendering from Few Images
               Using Transformers},
  booktitle = {Computer Vision -- ECCV 2022},
  pages     = {198--216},
  year      = {2022}
}

@inproceedings{gslrm,
  author    = {Zhang, Kai and Bi, Sai and Tan, Hao and Xiangli, Yuanbo and
               Zhao, Nanxuan and Sunkavalli, Kalyan and Xu, Zexiang},
  title     = {{GS-LRM}: Large Reconstruction Model for {3D} Gaussian
               Splatting},
  booktitle = {Computer Vision -- ECCV 2024},
  pages     = {1--19},
  year      = {2024}
}

@inproceedings{grm,
  author    = {Xu, Yinghao and Shi, Zifan and Wang, Yifan and
               Chen, Hansheng and Yang, Ceyuan and Peng, Sida and
               Shen, Yujun and Wetzstein, Gordon},
  title     = {{GRM}: Large Gaussian Reconstruction Model for Efficient
               {3D} Reconstruction and Generation},
  booktitle = {Computer Vision -- ECCV 2024},
  pages     = {1--20},
  year      = {2024}
}

@inproceedings{rayzer,
  author    = {Jiang, Hanwen and Tan, Hao and Wang, Peng and Jin, Haian and
               Zhao, Yue and Bi, Sai and Zhang, Kai and Luan, Fujun and
               Sunkavalli, Kalyan and Huang, Qixing and Pavlakos, Georgios},
  title     = {{RayZer}: A Self-Supervised Large View Synthesis Model},
  booktitle = {Proceedings of the IEEE/CVF International Conference on
               Computer Vision (ICCV)},
  pages     = {4918--4929},
  month     = oct,
  year      = {2025}
}

@inproceedings{efficientlvsm,
  author    = {Jia, Xiaosong and Sun, Yihang and You, Junqi and
               Wong, Songbur and Zou, Zichen and Yan, Junchi and
               Wu, Zuxuan and Jiang, Yu-Gang},
  title     = {{Efficient-LVSM}: Faster, Cheaper, and Better Large View
               Synthesis Model via Decoupled Co-Refinement Attention},
  booktitle = {The Fourteenth International Conference on Learning
               Representations},
  year      = {2026}
}

@inproceedings{avit,
  author    = {Yin, Hongxu and Vahdat, Arash and Alvarez, Jose M. and
               Mallya, Arun and Kautz, Jan and Molchanov, Pavlo},
  title     = {{A-ViT}: Adaptive Tokens for Efficient Vision Transformer},
  booktitle = {Proceedings of the IEEE/CVF Conference on Computer Vision
               and Pattern Recognition (CVPR)},
  pages     = {10809--10818},
  month     = jun,
  year      = {2022}
}

@inproceedings{ats,
  author    = {Fayyaz, Mohsen and Abbasi Koohpayegani, Soroush and
               Rezaei Jafari, Farnoush and Sengupta, Sunando and
               Vaezi Joze, Hamid Reza and Sommerlade, Eric and
               Pirsiavash, Hamed and Gall, J{\"u}rgen},
  title     = {Adaptive Token Sampling for Efficient Vision Transformers},
  booktitle = {Computer Vision -- ECCV 2022},
  pages     = {396--414},
  year      = {2022}
}

@inproceedings{park2026diversityawareviewpartitioningscalable,
  author    = {Park, Jinsoo and Choi, Donggyu and Seo, Ahyun and
               Cho, Minsu and Son, Jeany},
  title     = {Diversity-Aware View Partitioning for Scalable {VGGT}},
  booktitle = {Computer Vision -- ECCV 2026},
  pages     = {547--564},
  year      = {2026}
}

@inproceedings{settransformer,
  author    = {Lee, Juho and Lee, Yoonho and Kim, Jungtaek and
               Kosiorek, Adam and Choi, Seungjin and Teh, Yee Whye},
  title     = {{Set Transformer}: A Framework for Attention-based
               Permutation-Invariant Neural Networks},
  booktitle = {Proceedings of the 36th International Conference on
               Machine Learning},
  series    = {Proceedings of Machine Learning Research},
  volume    = {97},
  pages     = {3744--3753},
  publisher = {PMLR},
  year      = {2019}
}

@inproceedings{nair2025tokd,
  author    = {Nair, Nithin Gopalakrishnan and Kaza, Srinivas and
               Luo, Xuan and Patel, Vishal M. and Lombardi, Stephen and
               Park, Jungyeon},
  title     = {Scaling Transformer-Based Novel View Synthesis with Models
               Token Disentanglement and Synthetic Data},
  booktitle = {Proceedings of the IEEE/CVF International Conference on
               Computer Vision (ICCV)},
  pages     = {28567--28576},
  month     = oct,
  year      = {2025}
}

@article{kwak2026renov,
  author  = {Kwak, Min-Seop and Kwon, Minkyung and Choi, Jinhyeok and
             Park, Jiho and Kim, Seungryong},
  title   = {Projected Representation Conditioning for High-fidelity
             Novel View Synthesis},
  journal = {arXiv preprint arXiv:2602.12003},
  year    = {2026}
}

@inproceedings{wu2026projections,
  author    = {Wu, Zirui and Jiang, Zeren and Oswald, Martin R. and Song, Jie},
  title     = {From Rays to Projections: Better Inputs for
               Feed-Forward View Synthesis},
  booktitle = {Proceedings of the IEEE/CVF Conference on Computer Vision
               and Pattern Recognition (CVPR)},
  pages     = {29012--29022},
  month     = jun,
  year      = {2026}
}

@article{chen2026vggtdiff,
  author  = {Chen, Kangjie and Li, Xiangyu and Zhang, Dongbin and
             Zheng, Chaoda and Chen, Shijia and Deng, Jinhao and
             Lin, Hongbin and Choo, Sin Wai and Wang, Minqi and
             Yang, Minghao and Zhong, Dake and Song, Guorui and
             Zhang, Yu and Liu, Xianming and Wang, Boyang},
  title   = {{VGGT-Diff}: Visual Geometry Meets Diffusion for
             Sparse-View Novel View Synthesis},
  journal = {arXiv preprint arXiv:2609.33253},
  year    = {2026}
}

@inproceedings{Guo2024PRTGS,
  author    = {Guo, Yijia and Bai, Yuanxi and Hu, Liwen and Guo, Ziyi and
               Liu, Mianzhi and Cai, Yu and Huang, Tiejun and Ma, Lei},
  title     = {{PRTGS}: Precomputed Radiance Transfer of Gaussian Splats
               for Real-Time High-Quality Relighting},
  booktitle = {Proceedings of the 32nd ACM International Conference
               on Multimedia},
  pages     = {5112--5120},
  month     = oct,
  year      = {2024}
}

@inproceedings{Guo2024SpikeNeRF,
  author    = {Guo, Yijia and Bai, Yuanxi and Hu, Liwen and Liu, Mianzhi and
               Guo, Ziyi and Ma, Lei and Huang, Tiejun},
  title     = {{Spike-NeRF}: Neural Radiance Field Based On Spike Camera},
  booktitle = {2024 IEEE International Conference on Multimedia
               and Expo (ICME)},
  pages     = {1--6},
  month     = jul,
  year      = {2024}
}

@article{Guo2025MultiCameraRigs,
  author  = {Guo, Yijia and Hu, Tong and Li, Zhiwei and Hu, Liwen and
             Qian, Keming and Lin, Xitong and Chen, Shengbo and
             Huang, Tiejun and Ma, Lei},
  title   = {On-the-fly Large-scale {3D} Reconstruction from
             Multi-Camera Rigs},
  journal = {arXiv preprint arXiv:2512.08498},
  year    = {2025}
}
